\newif\ifIEEEFORMAT
\IEEEFORMATfalse

\newif\ifextendedversion
\extendedversiontrue
\ifIEEEFORMAT
  \documentclass[a4paper,10pt,conference]{ieeeconf}
  \IEEEoverridecommandlockouts
\else
  \documentclass[10pt,journal]{IEEEtran}
\fi

\usepackage[T1]{fontenc}
\usepackage[utf8]{inputenc}
\ifIEEEFORMAT
  \usepackage{amsmath,amssymb,amsfonts}

\else
  \usepackage{amsmath,amssymb,amsfonts}
\fi
\usepackage{newtxtext,newtxmath}
\usepackage{microtype}
\usepackage{graphicx}
\usepackage{booktabs}
\usepackage{tabularx}
\usepackage{array}
\usepackage{cite}
\usepackage[hidelinks]{hyperref}
\usepackage{xcolor}

\usepackage{enumitem}

\usepackage{tikz}
\usetikzlibrary{calc,positioning,fit}

\newcommand{\vrot}{v}
\newcommand{\arot}{\dot v}
\newcommand{\Pin}{P}
\newcommand{\Phat}{\widehat P}
\newcommand{\Jrot}{J}
\newcommand{\ctau}{c_\tau}
\newcommand{\ktl}{k_{\mathrm{tl}}}
\newcommand{\qlin}{q_1}
\newcommand{\taum}{\tau_{\mathrm m}}
\newcommand{\taud}{\tau_{\mathrm d}}
\newcommand{\Pone}{P_1}
\newcommand{\Ptwo}{P_2}
\newcommand{\Pthree}{P_3}
\newcommand{\Pfour}{P_4}
\newcommand{\Did}{\mathcal D_{\mathrm{id}}}
\newcommand{\Dval}{\mathcal D_{\mathrm{val}}}
\newcommand{\Eid}{\mathcal E_{\mathrm{id}}}
\newcommand{\Eval}{\mathcal E_{\mathrm{val}}}
\newcommand{\Aset}{\mathcal A}
\newcommand{\Aid}{\mathcal A_{\mathrm{id}}}
\newcommand{\Aval}{\mathcal A_{\mathrm{val}}}
\newcommand{\Pmin}{P_{\min}}

\newcommand{\figNestedPower}{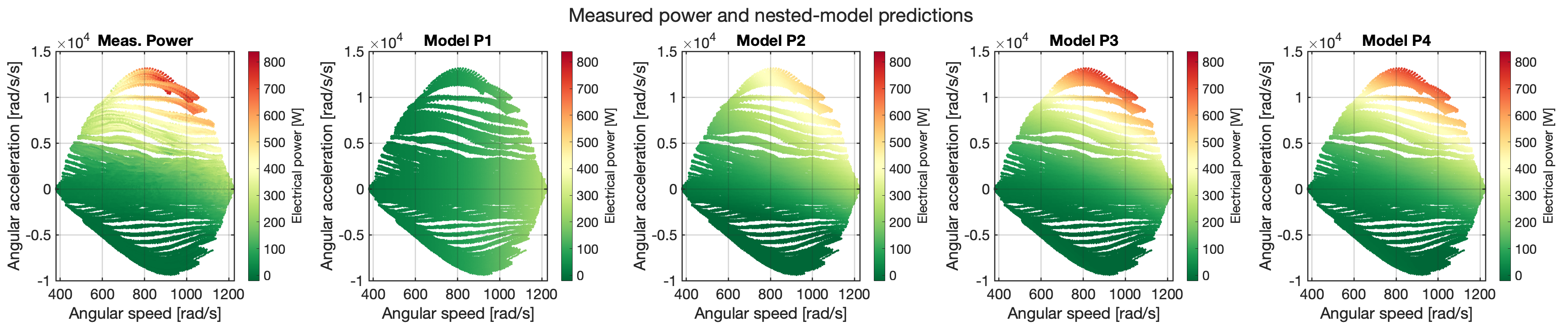}
\newcommand{\figNestedSummary}{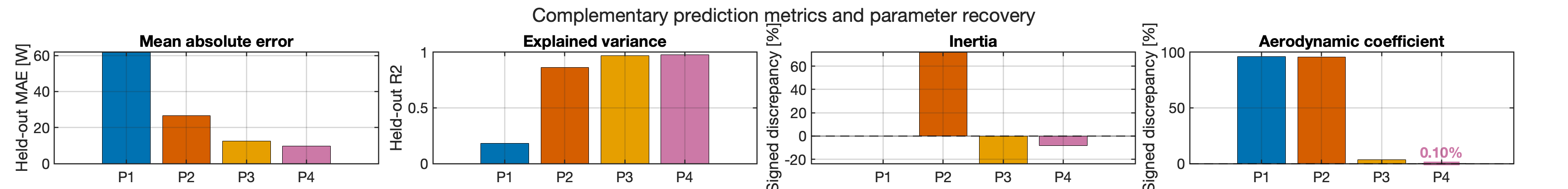}
\newcommand{\figNestedDiagnostics}{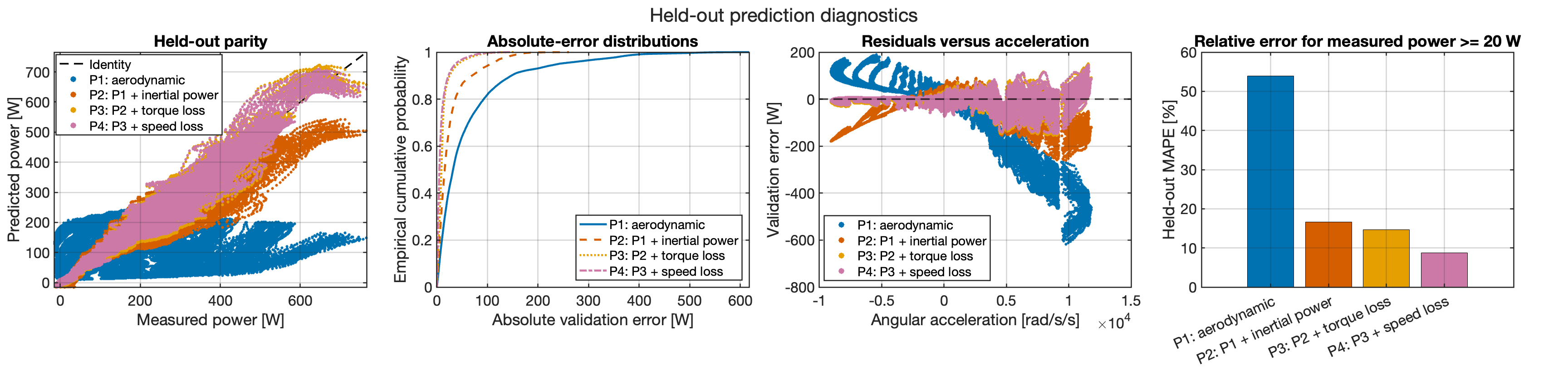}
\newcommand{\figNestedResiduals}{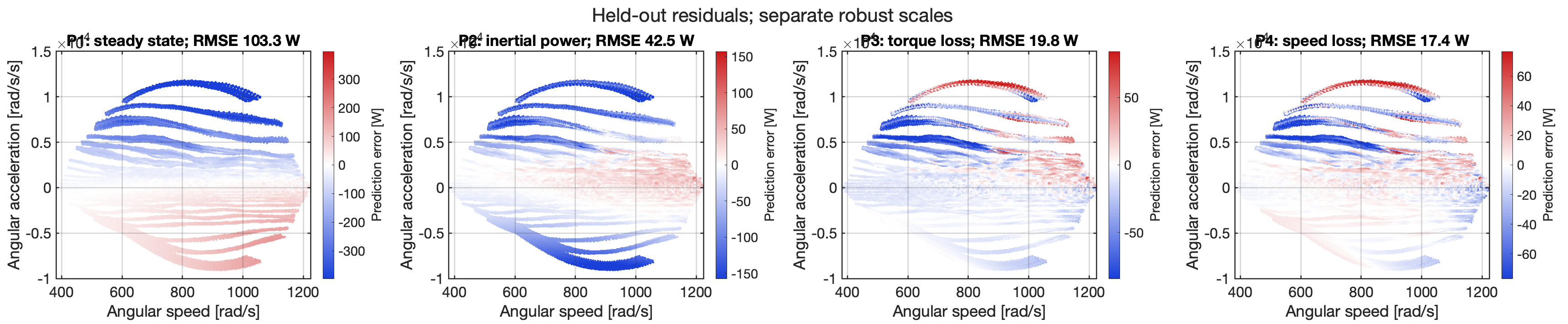}
\newcommand{\figNestedImprovements}{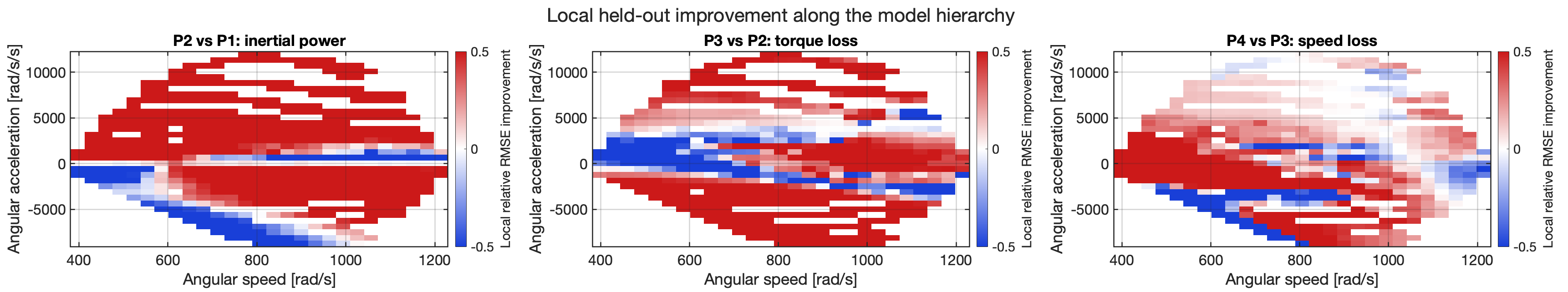}
\newcommand{\figNestedLosses}{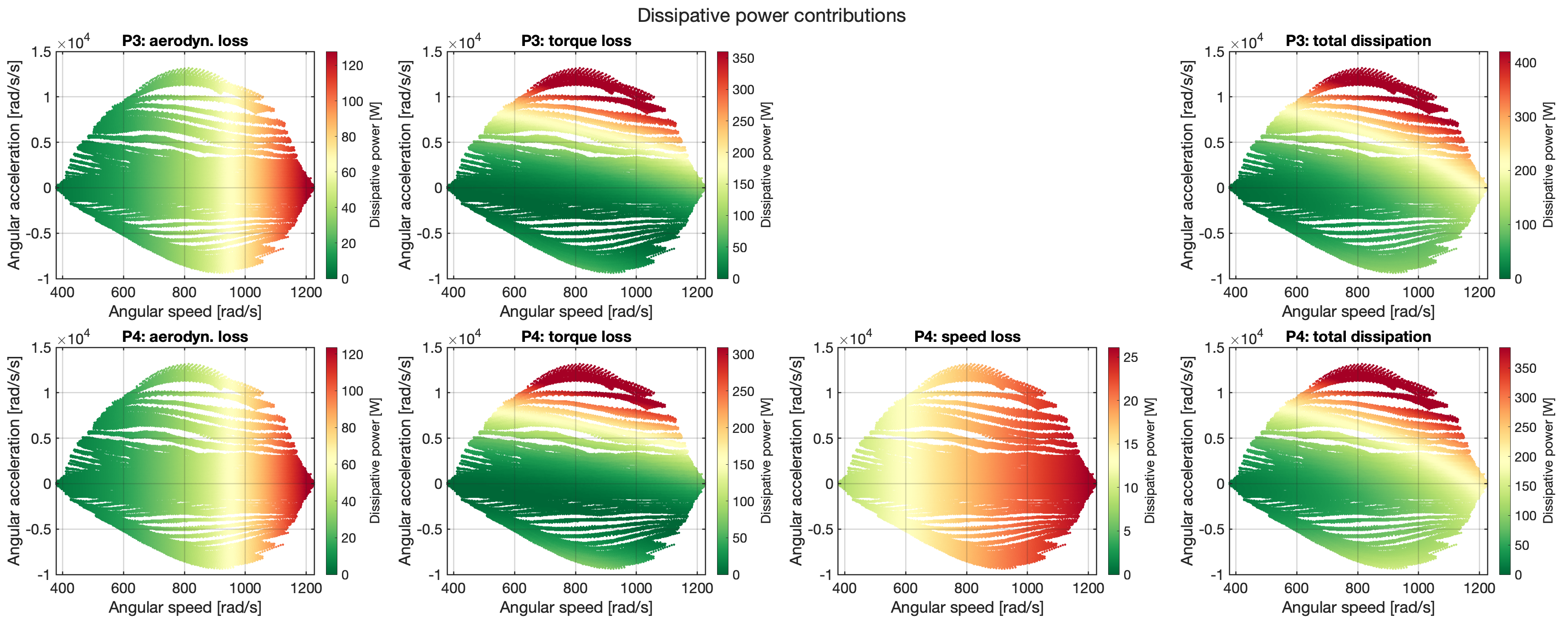}
\newcommand{\figCleaning}{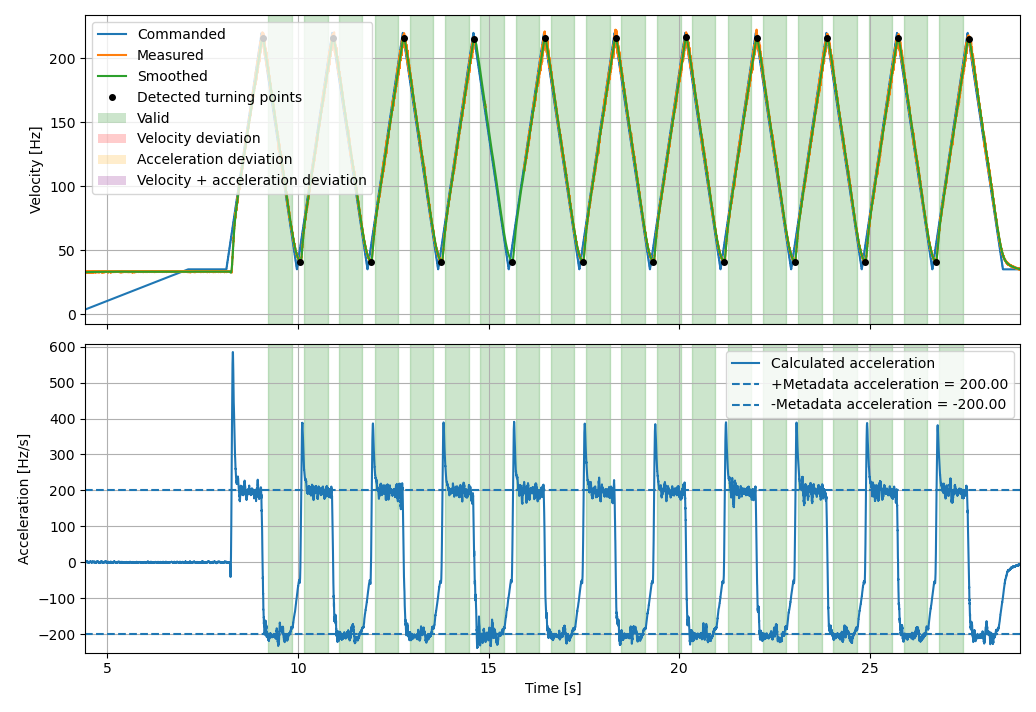}
\newcommand{\figRejected}{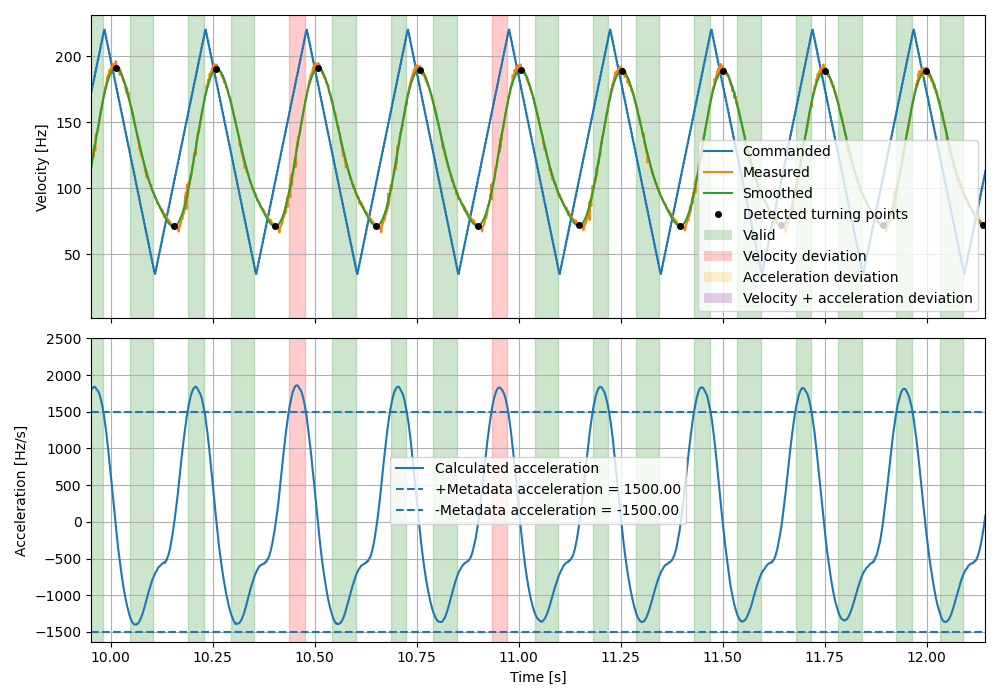}
\newcommand{\figHardwareA}{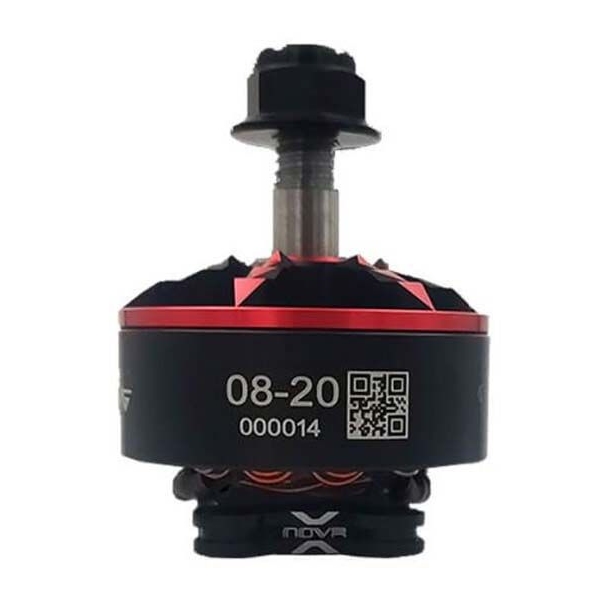}
\newcommand{\figHardwareB}{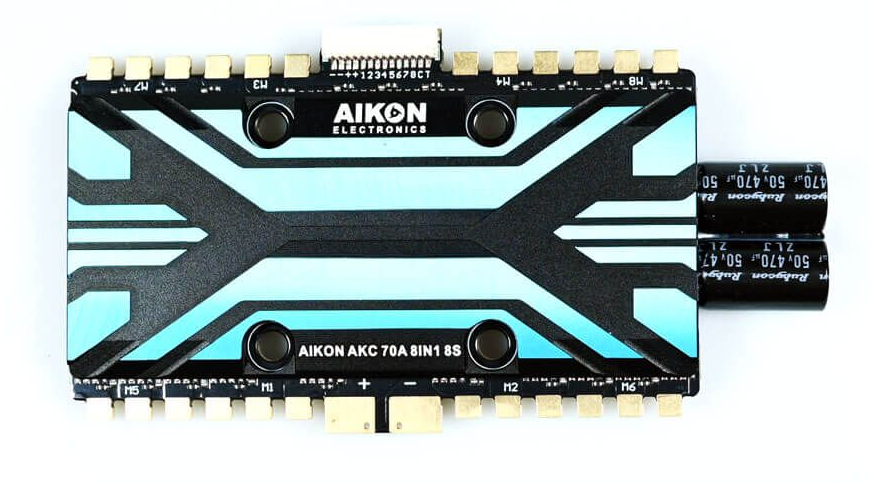}
\newcommand{\figHardwareC}{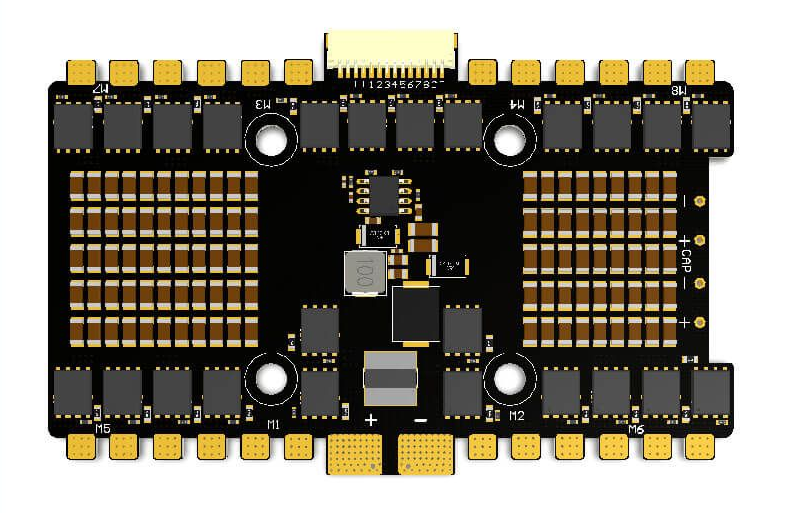}
\newcommand{\figHardwareD}{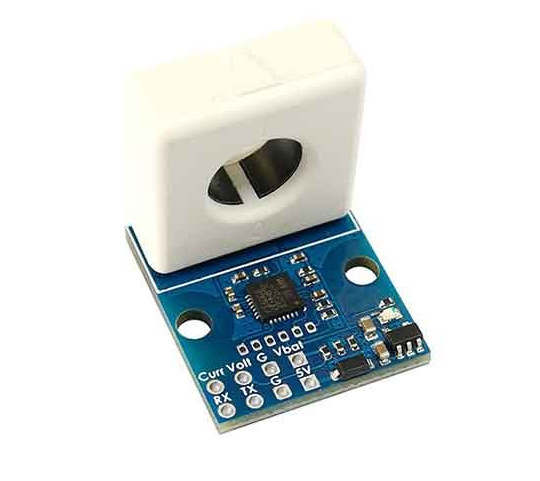}
\newcommand{\figHardwareE}{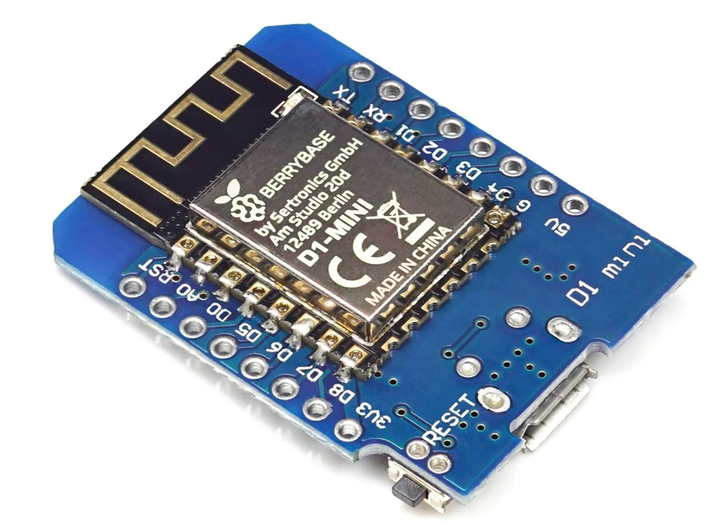}
\newcommand{\figHardwareF}{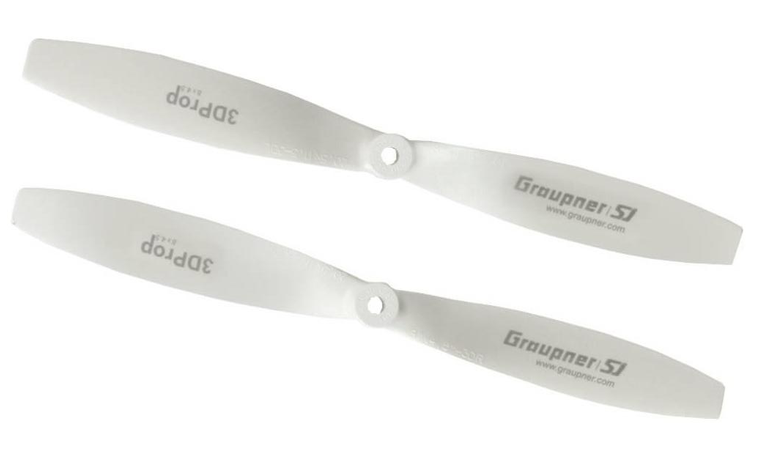}
\newcommand{\figInertiaSeparated}{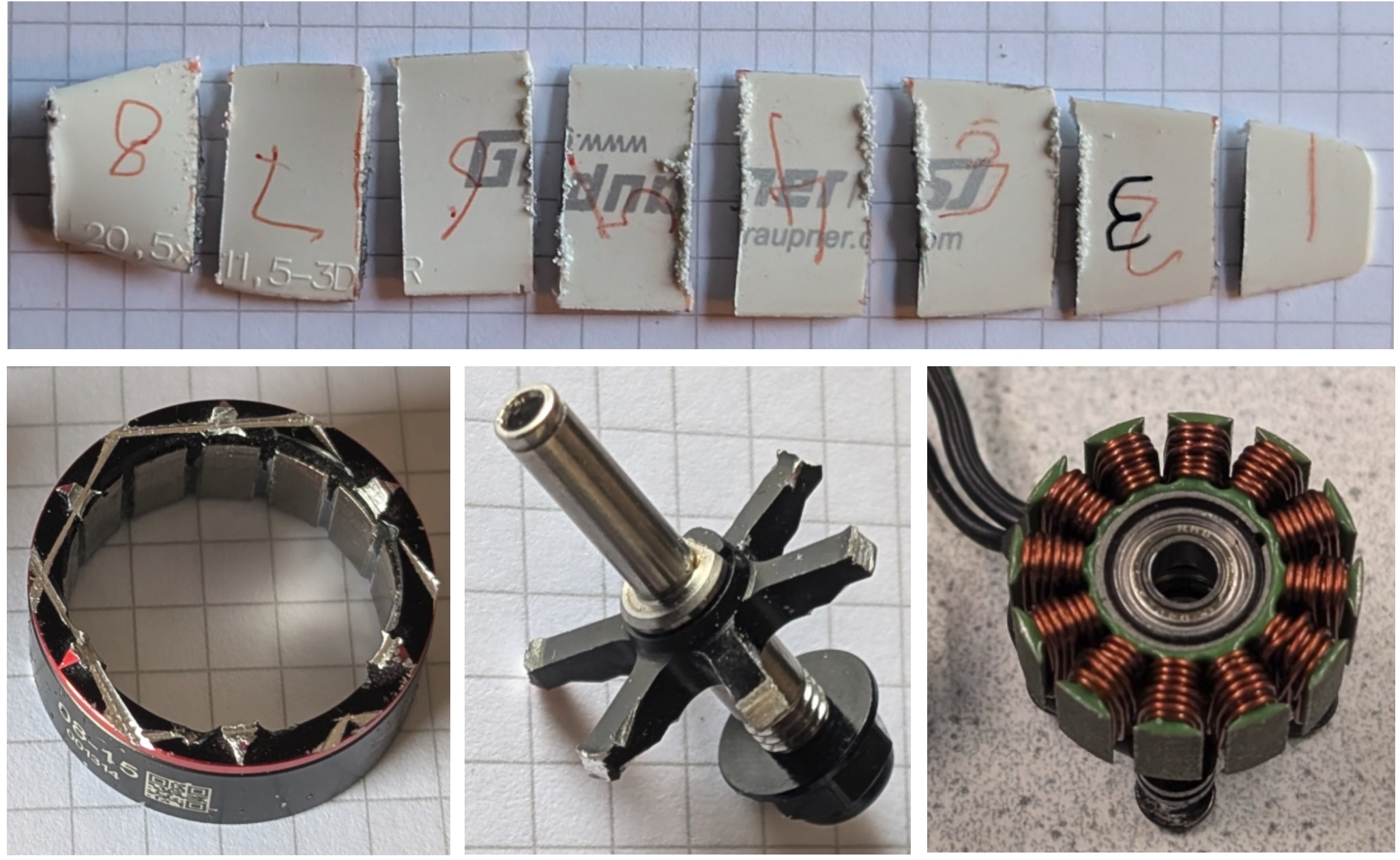}
\newcommand{\safegraphic}[2][]{\IfFileExists{#2}{\includegraphics[#1]{#2}}{\fbox{\parbox[c][0.12\textheight][c]{0.9\linewidth}{\centering Missing figure\\\texttt{\detokenize{#2}}}}}}

\title{Nested Power Models for Multirotor Propulsion: From Aerodynamic Drag to Electrical Losses}
\author{Antonio Franchi$^{1,2}$, 
Aaron Saini$^{1}$, Ahmed Ali$^{1}$, 
and Chiara Gabellieri$^{1}$, 
\thanks{$^1$Robotics and Mechatronics group, Faculty of Electrical Engineering, Mathematics and Computer Science, University of Twente, Enschede, The Netherlands. 
Email: schol@r-franchi.eu, c.gabellieri@utwente.nl\\
$^2$Department of Computer, Control and Management Engineering, Sapienza University of Rome, Rome, Italy. Email: schol@r-franchi.eu}%
\thanks{This work was partially funded by the Horizon Europe research project AUTOASSESS grant agreement No.~101120732 and by NWO OTP AVIATOR}}

\begin{document}
\maketitle

\begin{abstract}
Speed-only aerodynamic power models for multirotor propulsion cannot represent acceleration-dependent effects. This work develops a nested sequence of propulsion-power models that starts from aerodynamic power dissipation and progressively introduces a reversible kinetic-energy rate, torque-dependent electromechanical dissipation, and lumped speed-proportional dissipation. The models are identified using one subset of experiments and validated using the other on a motor-drive--propeller unit. Independent estimates of rotational inertia and aerodynamic drag complement predictive validation by assessing whether the models correctly attribute the measured power to reversible kinetic-energy exchange and irreversible dissipation and, within the latter, to aerodynamic and electromechanical losses. The results show that the reversible kinetic-energy rate is necessary but insufficient for accurate dynamic power prediction. Dissipation proportional to the squared motor torque provides the main additional improvement, while speed-proportional dissipation further prevents irreversible losses from being attributed to reversible kinetic-energy exchange. The resulting methodology provides a reusable and experimentally verifiable basis for developing and selecting dynamic propulsion-power models for multirotor systems.

\end{abstract}


\section{Introduction}
\label{sec:introduction}

The limited onboard energy of multirotor aerial vehicles makes propulsion-power modeling important for endurance estimation, vehicle and propulsion-system design, mission planning, trajectory generation, and control allocation. Flight-level models relate power consumption to vehicle velocity, thrust, climb rate, and flight direction to estimate consumption, range, endurance, and optimal flight speed \cite{Liu2017Power,Bauersfeld2022Range,Gong2023Power}. Experimental and data-driven approaches have similarly identified the power or energy consumption associated with complete flight trajectories \cite{AguilarLopez2022Energy}, while design-oriented methods support propulsion-system selection, sizing, and efficiency optimization \cite{Dai2019Efficiency,Kotarski2023HeavyLift,Joo2025Hybrid}. Together, these approaches demonstrate that compact and physically grounded power models can influence both vehicle design and the way aerial tasks are planned and executed. However, their primary objective is generally to predict total power, energy consumption, or propulsion performance.

At a finer level, propulsion models represent the battery, electronic speed controller, electric motor, and propeller as interconnected electrical, mechanical, and aerodynamic subsystems \cite{Coates2019Propulsion,Montoya2021Dynamic}. Such descriptions can predict thrust, rotor speed, current, or total electrical power and support component selection, diagnosis, simulation, and model-based control. They do not generally assess how the predicted power is attributed among reversible kinetic-energy exchange, aerodynamic dissipation, and electromechanical dissipation.

Dynamic power prediction is especially relevant when rotor speeds vary rapidly. Trajectory shape and acceleration can have a substantial effect on quadrotor energy consumption \cite{Alkomy2023Trajectories}, while multirotor control and experimental platforms increasingly support precise and agile motion \cite{Nascimento2019Survey,Baca2021MRS}. In these conditions, the power required to accelerate the motor--propeller assembly is not described by a steady-state speed-only relation. Rotor acceleration changes the rotational kinetic energy, but it also changes motor torque and current and may therefore affect irreversible electromechanical dissipation.

Furthermore, rapid rotor-speed modulation is not limited to fast translational flight. It also arises in aerial physical interaction, where the vehicle may remain nearly stationary with respect to the surrounding air while producing rapidly varying forces and torques against the environment. Aerial manipulators and multidirectional-thrust platforms have been developed for contact inspection, manipulation, and other interaction tasks \cite{Ruggiero2018Review,Ollero2022Past}. Fully actuated and redundantly actuated multirotors can regulate interaction wrenches and vehicle attitude with limited or no translational motion, but doing so relies directly on coordinated and potentially rapid variations of individual rotor speeds \cite{Ryll2017Interaction,Tognon2019Redundant,Dyer2019Energy}. Related systems employ reconfigurable propulsion or contact mechanisms to improve interaction accuracy \cite{Taylor2024Reconfigurable}, while recent omnidirectional aerial-interaction experiments further demonstrate the relevance of rotor-level actuation models during contact-rich operation \cite{Veenstra2026Interaction}. More generally, the growing range of multirotor physical-interaction tasks makes actuator dynamics, limits, and energetic behavior increasingly important \cite{Xu2023PhysicalSurvey}.

Despite these developments, accurate total-power prediction does not establish correct attribution among the underlying physical contributions. During accelerated rotor motion, reversible kinetic-energy exchange, aerodynamic dissipation, and torque-dependent electromechanical losses may have correlated effects over a finite operating domain. A model may therefore reproduce the measured electrical power while assigning irreversible loss to the reversible kinetic-energy rate or compensating an inaccurate aerodynamic contribution through an electromechanical-loss term. Such confounding becomes problematic when the model must determine not only the energy required by a trajectory, but also how much is temporarily stored, how much is irreversibly lost, and which propulsion component should be improved.

This work addresses this gap by developing and experimentally assessing propulsion-power models with progressively richer energetic decompositions. The models are identified on one group of complete dynamic experiments and evaluated on separate experiments excluded from identification. Held-out predictive validation is complemented by independent estimates of rotational inertia and aerodynamic drag to assess whether the measured power is attributed consistently among reversible kinetic-energy exchange, aerodynamic dissipation, and electromechanical dissipation.

The main contribution of this work is a physically motivated sequence of multirotor propulsion-power models that starts from aerodynamic power dissipation and progressively introduces the reversible kinetic-energy rate, torque-dependent electromechanical dissipation, and lumped speed-proportional dissipation. Experiments show that the kinetic-energy rate is necessary but insufficient during accelerated rotor motion, that torque-dependent dissipation provides the principal additional predictive improvement, and that speed-proportional dissipation further improves power attribution and the joint recovery of inertia and aerodynamic drag.

This contribution is supported by:
\begin{enumerate}[leftmargin=*]
\item a dynamic identification methodology based on realizable speed--acceleration trajectories and validation on complete experiments conducted at nominal acceleration rates excluded from identification; and

\item a validation framework combining held-out predictive accuracy with independent inertia and aerodynamic-drag estimates to assess the attribution of measured power among reversible kinetic-energy exchange, aerodynamic dissipation, and electromechanical dissipation.
\end{enumerate}

The remainder of this article is organized as follows.
\ifextendedversion Section~\ref{sec:related-work} reviews power and energy models across aerial-vehicle configurations and modeling scales, and positions the contribution of this work. \fi
Section~\ref{sec:models} develops the nested propulsion-power models. Section~\ref{sec:method} defines the criteria and methodology used to identify and assess them, while Section~\ref{sec:implementation} presents the experimental realization and independent aeromechanical estimates. Section~\ref{sec:results} evaluates predictive accuracy and power attribution along the model sequence, and Section~\ref{sec:discussion} discusses the scope, implications, and conclusions.

\ifextendedversion
\section{Related Work}
\label{sec:related-work}

Power and energy models for aerial vehicles differ substantially in physical resolution and intended use. Component-level models describe propeller loads, motor torque, current, voltage, conversion losses, and battery behavior. Vehicle-level models aggregate propulsion, airframe aerodynamics, and flight dynamics to estimate range, endurance, or mission energy. Planning and control methods then use these models as objectives or constraints for trajectory generation, mission scheduling, and control allocation. Because rotary-wing, fixed-wing, flapping-wing, convertible, and physically interactive aircraft emphasize different energetic mechanisms, this section follows the modeling chain from propulsion fundamentals to vehicle- and mission-level optimization, then positions the rotor-level attribution problem addressed in this work.

\subsection{Rotor aerodynamics and speed-only power models}

The cubic dependence of rotor power on angular speed originates from standard rotor and propeller aerodynamics. For fixed geometry and operating conditions, thrust and resisting torque are commonly approximated as quadratic functions of rotor speed, so mechanical power scales cubically. Low-Reynolds-number propeller measurements and blade-element--momentum models show, however, that thrust and power coefficients depend on propeller geometry, inflow, Reynolds number, and scale effects \cite{Brandt2011Propeller,McCrink2017BladeElement}. Multirotor models consequently range from constant-coefficient quadratic thrust and drag-torque laws used in dynamics and control \cite{Mahony2012Multirotor} to rotor-power-coefficient models embedded in flight simulation \cite{Davoudi2020QuadRotor}.

When the aerodynamic coefficients are treated as constant, these actuator laws induce the speed-only cubic power model. This representation has supported aerodynamic-power control at the propulsion-unit level \cite{Bangura2014Aerodynamic}, system-level power-consumption models \cite{Liu2017Power}, and efficiency analysis of morphing multirotors \cite{Aboudorra2024OmniMorph}. Such models compactly represent aerodynamic dissipation, but speed alone cannot express the reversible kinetic-energy exchange and the additional losses associated with rotor acceleration that are central to the present work.

\subsection{Electric propulsion-chain modeling and design}

A second line of work models the propulsion chain by interconnecting the power source, electronic speed controller, motor, shaft, and propeller. Analytical design methods combine propeller theory, motor constants, current and speed limits, battery capacity, and vehicle requirements to select components and estimate performance \cite{Mohamed2011Designing,Gatti2017Complete}. Broader sizing correlations extend this reasoning across electric and fuel-powered UAV categories \cite{Alulema2021Sizing}, while multicopter-specific optimization selects motor, propeller, and battery combinations for efficiency \cite{Dai2019Efficiency}. Experimental characterization and sizing studies further address large or heavy-lift multirotors whose propulsion components operate far from the small-quadrotor regime \cite{Kotarski2023HeavyLift}.

Component-resolved dynamic models have been developed for both fixed-wing and multirotor propulsion. Fixed-wing formulations separately represent the electrical source, controller, motor, and propeller to predict thrust, rotor speed, and current \cite{Coates2019Propulsion}; related dynamic models describe the coupled electric motor and propeller over transient operation \cite{Montoya2021Dynamic}. More recent physics-informed approaches combine dimensional analysis, propeller data, test-rig measurements, and torque-based current prediction to improve cross-configuration endurance estimation \cite{Dai2026HighPrecision}. These approaches increase propulsion-model fidelity, whereas the present study asks which compact acceleration-dependent terms are identifiable from source-side power and whether the resulting decomposition is physically consistent.

\subsection{Battery, range, endurance, and vehicle-level power}

Battery-aware aircraft models connect propulsion demand to stored electrical energy, voltage variation, discharge rate, and degradation. Classical electric-aircraft analyses derive range and endurance while accounting for battery-discharge behavior \cite{Traub2011Range}; subsequent work examines how battery degradation changes optimal endurance and range over the aircraft life cycle \cite{Paek2020Battery}. For multirotors, practical endurance estimators combine hover and forward-flight requirements with battery and payload information \cite{Hwang2018Endurance}, while integrated aerodynamic, motor, and gray-box battery models estimate range, endurance, and optimal speed across multicopter designs \cite{Bauersfeld2022Range}. Alternative architectures seek endurance gains through hybrid power systems \cite{Joo2025Hybrid} or solar augmentation at MAV scale \cite{Hassanalian2014Solar}.

At vehicle level, empirical models fit power or energy directly from flight measurements and operating conditions. Examples include comprehensive battery-based energy models for UAV maneuvers \cite{Abeywickrama2018Comprehensive}, quadrotor energy models identified from experimental voltage data \cite{AguilarLopez2022Energy}, and comparisons of drone and ground-vehicle delivery energy \cite{Kirschstein2020Delivery}. Analytical multirotor models instead express power as a function of thrust, forward velocity, climb rate, flight direction, and aerodynamic contributions \cite{Gong2023Power}. Reviews of drone-energy factors and models show that the appropriate abstraction depends strongly on whether the objective is component design, flight-time estimation, mission planning, or energy management \cite{Beigi2022Overview}. Across these approaches, accurate total-energy prediction does not by itself establish whether transient power is attributed correctly among stored energy and individual loss mechanisms.

\subsection{Energy-aware trajectory, allocation, and mission planning}

Power models become decision criteria when trajectories, actuator commands, or missions are optimized. Polynomial-trajectory studies show that kinematic choices can substantially change quadrotor energy consumption \cite{Alkomy2023Trajectories}. Integrated physical models can be embedded directly in trajectory optimization and feedback control, yielding experimentally verified reductions in consumed energy \cite{Michel2023EnergyOptimal}. Coverage-planning methods use experimentally derived energy models together with sensing-resolution and return-safety constraints \cite{DiFranco2016Coverage}, while multi-UAV planners jointly optimize coverage assignment and flight speed under battery constraints \cite{Datsko2024Coverage}. Recent logistics planners further combine multirotor dynamics, obstacle avoidance, mission duration, payload, and energy-efficiency objectives \cite{Wu2025PathPlanning}.

Communication and deployment studies use rotary-wing propulsion models to optimize trajectory and service time jointly with communication energy \cite{Zeng2019Energy}. Comparative mission simulators extend this perspective to both rotary-wing and fixed-wing vehicles \cite{Ribeiro2024RotaryFixed}. Within redundant actuation, energy-optimal allocation selects among multiple actuator combinations that realize the same body wrench \cite{Dyer2019Energy}. These applications make model attribution consequential: if a transient reversible term is treated as dissipation, or an irreversible loss is hidden in an inertial coefficient, the optimization may select a different trajectory, speed, or actuator evolution than a physically resolved model would.

\subsection{Fixed-wing, flapping-wing, and convertible MAVs}

Fixed-wing and flapping-wing MAVs emphasize energetic mechanisms that differ from hovering multirotors. Long-endurance fixed-wing MAV design couples aerodynamic efficiency, lightweight construction, propulsion selection, speed, and stored energy \cite{Bronz2009Endurance}. The program in  \cite{DeCroon2009DelFly} integrates aerodynamic, mechanical, sensing, and autonomy constraints in flapping-wing systems, while subsequent robotic-flapper experiments reveal unsteady aerodynamic coupling during rapid maneuvers \cite{Karasek2018Flapper}. These platforms reinforce the broader need to separate actuator-energy storage, aerodynamic force production, and irreversible conversion losses even though their actuation principles differ from those considered here.

Convertible aircraft combine multirotor-like vertical flight with wing-borne cruise. Tailsitter control spans hover, transition, and forward flight despite strongly changing and imperfectly known aerodynamics \cite{Smeur2020Incremental}; tilt-rotor VTOL systems add configuration redundancy and dynamic allocation under actuator failures \cite{Mousaei2022TiltRotor}. Reconfigurable and morphing multirotors likewise use additional actuation to change the relation between rotor thrust and body wrench \cite{Taylor2024Reconfigurable}. At the vehicle-design level, a fully actuated tilting-propeller quadrotor demonstrates the modeling and control implications of redundant actuation \cite{Ryll2015Overactuated}. These systems motivate energetic models that remain meaningful when rotor states change rapidly even if the vehicle motion changes comparatively little.

\subsection{Aerodynamic interaction and physically interactive flight}

The surrounding flow can change rotor thrust, resisting torque, and efficiency. Ground-effect experiments characterize both complete and partial ground interaction in multirotors \cite{SanchezCuevas2017Ground}, while obstacle-proximity studies map aerodynamic effects near surfaces and infrastructure \cite{SanchezCuevas2020Obstacles}. Rotor--rotor interference, vehicle-body interaction, payload aerodynamics, and wake recirculation create further deviations from isolated-rotor models. Such effects are especially relevant in aerial manipulation and inspection, where the vehicle operates close to the environment and may sustain interaction forces with limited translational motion.

The aerial-manipulation literature ranges from early system \cite{Ruggiero2018Review} to broad assessments of the field's development and open challenges \cite{Ollero2022Past,Xu2023PhysicalSurvey}. Fully actuated and redundantly actuated platforms support six-dimensional interaction and contact-force regulation \cite{Ryll2017Interaction,Tognon2019Redundant}, while recent reconfigurable and omnidirectional systems extend these capabilities \cite{Veenstra2026Interaction}. Control surveys and shared experimental infrastructures have helped standardize multirotor modeling, control, and reproducible deployment \cite{Nascimento2019Survey,Baca2021MRS}. In these interaction regimes, rotor acceleration and source power can vary substantially even when vehicle velocity is small, making rotor-level transient-power models particularly relevant.

\subsection{Position of the present work}

The reviewed literature provides models for propeller aerodynamics, electric propulsion, battery discharge, vehicle-level consumption, endurance, trajectory optimization, convertible flight, redundant allocation, and physical interaction. These models have enabled increasingly accurate prediction and optimization across diverse aerial platforms. Their principal outputs are typically thrust, current, total power, range, endurance, or mission energy. Comparatively less attention has been given to experimentally determining which acceleration-dependent terms are required to attribute source-side electrical power consistently among reversible kinetic-energy exchange, aerodynamic dissipation, and electromechanical dissipation. The present work addresses this attribution problem through a nested model sequence, held-out dynamic experiments, and independent aeromechanical reference estimates.
\fi

\section{Nested Propulsion-Power Models}\label{sec:models}

This section describes the nested sequence of propulsion-power models developed in this work to progressively represent aerodynamic, inertial, and electromechanical effects.
The ordering follows the representational limitations of the models: each model introduces a contribution that its predecessor cannot represent, namely acceleration-dependent reversible power, acceleration-dependent dissipation, and finally distinct speed-proportional dissipation.

Consider a motor with an electronic speed controller driving a rigidly connected propeller in one rotational direction. Let $\vrot\geq0$ denote the rotor angular speed and $\arot$ its angular acceleration. The total moment of inertia of the rotating assembly is the constant $\Jrot>0$. Under quasi-steady aerodynamics, the resisting torque associated with the so-called drag moment is
\begin{equation}
\taud(\vrot)=\ctau\vrot^2, \quad \ctau>0.
\label{eq:drag-torque}
\end{equation}

The conventional \emph{one-parameter speed-only aerodynamic model}, used for instance in the multirotor power analysis of \cite{Aboudorra2024OmniMorph}, approximates the power required to spin a rotor as
\begin{equation}
\Pone(\vrot):=\ctau|\vrot|^3
=\vrot\taud(\vrot),
\qquad \vrot\geq0.
\label{eq:p1}
\end{equation}
The absolute value, although redundant over the investigated domain, makes the dissipative and thus nonnegative aerodynamic power explicit. The model depends only on speed and is often motivated by aerodynamic dissipation being dominant at typical operating speeds. Its advantage is simplicity, whereas its limitations are that it cannot represent acceleration-dependent power and attributes all predicted power to irreversible aerodynamic dissipation. Consequently, when identified from accelerated-motion data, its aerodynamic term may absorb reversible kinetic-energy exchange, thereby conflating recoverable energy with irreversible loss.

To represent the reversible contribution explicitly, consider the aeromechanical motor torque required to drive the propeller at a given $(\vrot,\arot)$:
\begin{equation}
\taum:=\Jrot\arot+\taud(\vrot)
=\Jrot\arot+\ctau\vrot^2.
\label{eq:motor-torque}
\end{equation}
The first term changes the rotational kinetic energy, whereas the second overcomes aerodynamic drag. The corresponding mechanical power defines the \emph{two-parameter dynamic mechanical-power model}:
\begin{equation}
\Ptwo(\vrot,\arot)
:=\vrot\taum
=\Jrot\vrot\arot+\ctau|\vrot|^3
=\Pone(\vrot)+\Jrot\vrot\arot.
\label{eq:p2}
\end{equation}
The inertial contribution satisfies $\Jrot\vrot\arot=\mathrm{d}(\Jrot\vrot^2/2)/\mathrm{d}t$ and is reversible, whereas the aerodynamic contribution is nonnegative and irreversible. In bond-graph or power-port terminology, the former exchanges power with a kinetic-energy storage element, while the latter represents an aerodynamic dissipation port.

Model $\Ptwo$ thus distinguishes acceleration-dependent power exchanged with kinetic-energy storage from aerodynamic dissipation. However, because its dissipative contribution depends only on speed, acceleration-dependent losses may be attributed to the reversible kinetic-energy-rate term, compromising the separation between stored and dissipated power.

\begin{table*}[t]
\centering
\caption{Nested electrical-power models and their energetic structure.}
\label{tab:nested-models}

\renewcommand{\arraystretch}{1.25}
\setlength{\tabcolsep}{4pt}
\begin{tabularx}{\textwidth}{
    >{\raggedright\arraybackslash}p{0.20\textwidth}
    >{\raggedright\arraybackslash}p{0.14\textwidth}
    >{\raggedright\arraybackslash}p{0.18\textwidth}
    >{\raggedright\arraybackslash}p{0.22\textwidth}
    >{\raggedright\arraybackslash}X}
\toprule
\textbf{Nested model}
&
\textbf{Model parameters and input variables}
&
\textbf{Reversible power contribution}
&
\textbf{Irreversible power loss}
&
\textbf{Parameters and variables contributing to dissipation}
\\
\midrule

$\Pone(\vrot)=\ctau|\vrot|^3$
&
$\ctau$; $\vrot$
&
Absent
&
Aerodynamic:
$\ctau|\vrot|^3$
&
$\ctau$, $\vrot$
\\

$\Ptwo(\vrot,\arot)
=\Pone(\vrot)+\Jrot\vrot\arot$
&
$\ctau$, $\Jrot$; $\vrot$, $\arot$
&
Kinetic-energy rate:
$\Jrot\vrot\arot$
&
Aerodynamic:
$\ctau|\vrot|^3$
&
$\ctau$, $\vrot$
\\

$\Pthree(\vrot,\arot)
=\Ptwo(\vrot,\arot)+\ktl\taum^2$
&
$\ctau$, $\Jrot$, $\ktl$; $\vrot$, $\arot$
&
Kinetic-energy rate:
$\Jrot\vrot\arot$
&
Aerodynamic and electromechanical:
$\ctau|\vrot|^3+\ktl\taum^2$
&
$\ctau$, $\Jrot$, $\ktl$;
$\vrot$, $\arot$
\\

$\Pfour(\vrot,\arot)
=\Pthree(\vrot,\arot)+\qlin|\vrot|$
&
$\ctau$, $\Jrot$, $\ktl$, $\qlin$;
$\vrot$, $\arot$
&
Kinetic-energy rate:
$\Jrot\vrot\arot$
&
Aerodynamic, electromechanical, and lumped:
$\ctau|\vrot|^3+\ktl\taum^2+\qlin|\vrot|$
&
$\ctau$, $\Jrot$, $\ktl$, $\qlin$;
$\vrot$, $\arot$
\\

\bottomrule
\end{tabularx}
\end{table*}

\begin{figure}[t]
\centering

\resizebox{\columnwidth}{!}{%
\begin{tikzpicture}[
    font=\small,
    >=latex,
    box/.style={
        draw,
        rounded corners=2pt,
        align=center,
        line width=0.8pt,
        inner xsep=2.5pt,
        inner ysep=2.5pt
    },
    source/.style={
        box,
        text width=27mm,
        minimum height=19mm,
        fill=blue!7,
        draw=blue!55!black
    },
    interconnection/.style={
        box,
        text width=59mm,
        minimum height=19mm,
        fill=gray!8,
        draw=black!70,
        line width=1pt
    },
    contribution/.style={
        box,
        text width=24mm,
        minimum height=27mm
    },
    aero/.style={
        contribution,
        fill=orange!10,
        draw=orange!65!black
    },
    storage/.style={
        contribution,
        fill=green!8,
        draw=green!45!black
    },
    electromech/.style={
        contribution,
        fill=red!7,
        draw=red!55!black
    },
    lumped/.style={
        contribution,
        fill=violet!7,
        draw=violet!55!black
    },
    modeltag/.style={
        rounded corners=1.5pt,
        inner xsep=2.5pt,
        inner ysep=1pt,
        font=\footnotesize\bfseries,
        text=white
    },
    powerbond/.style={
        ->,
        >=latex,
        line width=1.05pt
    },
    reversiblebond/.style={
        <->,
        >=latex,
        line width=1.05pt
    },
    bondlabel/.style={
    font=\footnotesize,
    align=center,
    inner xsep=2pt,
    inner ysep=1.5pt,
    rounded corners=1pt,
    fill=gray!12,
    draw=none,
    text=black!75
}
]

\node[source] (source) at (-3.8,3.25) {
    \textbf{Electrical source port}\\[0.6mm]
    $(V_{\mathrm{dc}},I_{\mathrm{dc}})$\\[0.6mm]
    $\Pin=V_{\mathrm{dc}}I_{\mathrm{dc}}$
};

\node[interconnection] (dirac) at (2.3,3.25) {
    \textbf{Power-preserving}\\[-0.2mm]
    \textbf{interconnection}\\
    \footnotesize Dirac-structure layer\\[0.6mm]
    $(V_{\mathrm{dc}},I_{\mathrm{dc}})
    \longleftrightarrow
    (\taum,\vrot)$\\[0.6mm]
    \footnotesize power routing without storage or dissipation
};

\draw[reversiblebond, blue!45!black]
    (source.east) --
    node[
        bondlabel,
        above,
        yshift=-4mm
    ] {
        source\\
        power\\[1.5mm]
        $\Pin$
    }
    (dirac.west);

\coordinate (diracPone)
    at ($(dirac.south west)!0.05!(dirac.south east)$);

\coordinate (diracPtwo)
    at ($(dirac.south west)!0.375!(dirac.south east)$);

\coordinate (diracPthree)
    at ($(dirac.south west)!0.625!(dirac.south east)$);

\coordinate (diracPfour)
    at ($(dirac.south west)!0.95!(dirac.south east)$);

\node[aero] (raero) at (-4.05,-1.05) {
    \vphantom{\rule{0pt}{3mm}}\\[-1mm]
    \textbf{Aerodynamic}\\[-0.2mm]
    \textbf{dissipation}\\[0.7mm]
    $\taud(\vrot)=\ctau\vrot^2$\\[0.7mm]
    $P_{\mathrm{aero}}=\ctau|\vrot|^3$
};

\node[storage] (inertia) at (-1.35,-1.05) {
    \vphantom{\rule{0pt}{3mm}}\\[-1mm]
    \textbf{Kinetic-energy}\\[-0.2mm]
    \textbf{storage}\\[0.7mm]
    $H(\vrot)=\tfrac{1}{2}\Jrot\vrot^2$\\[0.7mm]
    $\dot H=\Jrot\vrot\arot$
};

\node[electromech] (rtl) at (1.35,-1.05) {
    \vphantom{\rule{0pt}{3mm}}\\[-1mm]
    \textbf{Torque-dependent}\\[-0.2mm]
    \textbf{electromechanical}\\[-0.2mm]
    \textbf{dissipation}\\[0.7mm]
    $\taum=\Jrot\arot+\ctau\vrot^2$\\[0.7mm]
    $P_{\mathrm{tl}}=\ktl\taum^2$
};

\node[lumped] (rlin) at (4.05,-1.05) {
    \vphantom{\rule{0pt}{3mm}}\\[-1mm]
    \textbf{Lumped}\\[-0.2mm]
    \textbf{speed-proportional}\\[-0.2mm]
    \textbf{dissipation}\\[0.7mm]
    $P_{\mathrm{lin}}=\qlin|\vrot|$
};

\draw[powerbond, orange!70!black]
    (diracPone)
    to[out=-105,in=90]
    (raero.north);

\draw[reversiblebond, green!45!black]
    (diracPtwo)
    to[out=-100,in=90]
    (inertia.north);

\draw[powerbond, red!60!black]
    (diracPthree)
    to[out=-85,in=90]
    (rtl.north);

\draw[powerbond, violet!60!black]
    (diracPfour)
    to[out=-75,in=90]
    (rlin.north);

\node[
    bondlabel,
    text width=16mm
] (labelPone) at (-2.35,1.45) {
    aerodynamic irreversible
    power $P_{\mathrm{aero}}$
};

\node[
    bondlabel,
    text width=14mm
] (labelPtwo) at (0.15,1.25) {
    reversible
    power
    $\dot H$
};

\node[
    bondlabel,
    text width=15mm
] (labelPthree) at (2.2,1.25) {
    electromech.
    irreversible power $P_{\mathrm{tl}}$
};

\node[
    bondlabel,
    text width=16mm
] (labelPfour) at (4.15,1.25) {
    lumped irreversible
    power $P_{\mathrm{lin}}$
};

\node[
    modeltag,
    fill=orange!75!black,
    anchor=north east
] at ($(raero.north east)+(-1mm,-1mm)$)
    {$\Pone$, $\Ptwo$, $\Pthree$, $\Pfour$};

\node[
    modeltag,
    fill=green!50!black,
    anchor=north east
] at ($(inertia.north east)+(-1mm,-1mm)$)
    {$\Ptwo$, $\Pthree$, $\Pfour$};

\node[
    modeltag,
    fill=red!65!black,
    anchor=north east
] at ($(rtl.north east)+(-1mm,-1mm)$)
    {$\Pthree$,$\Pfour$};

\node[
    modeltag,
    fill=violet!65!black,
    anchor=north east
] at ($(rlin.north east)+(-1mm,-1mm)$)
    {$\Pfour$};

\end{tikzpicture}%
}

\caption{Port-Hamiltonian-inspired representation of the energetic structure underlying the nested models. The electrical source supplies $\Pin$ through a conceptual power-preserving interconnection. The kinetic-energy storage element exchanges reversible power, whereas the other branches represent aerodynamic, torque-dependent electromechanical, and lumped speed-proportional dissipation. Accordingly, each model combines the contributions indicated by its labels. The diagram provides an energetic interpretation rather than a complete bond graph of the motor-drive--propeller system.}
\label{fig:energetic-architecture}
\end{figure}
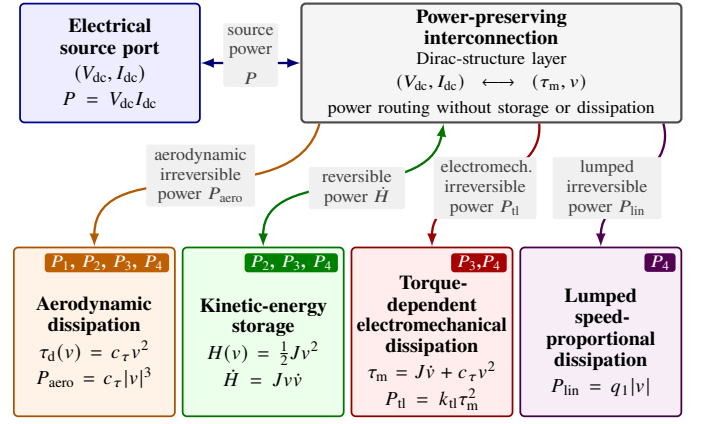

\smallskip
Because a speed-only dissipative correction would not address this missing acceleration dependence, the next extension introduces acceleration-dependent dissipation through the motor torque. Denote by $I_{\mathrm m}$ the motor current. Under the ideal torque-current relation $\taum=k_\tau I_{\mathrm m}$, where $k_\tau>0$ is the motor torque constant, the winding copper loss is proportional to $I_{\mathrm m}^2$ and, therefore, to $\taum^2$. This motivates the following \emph{three-parameter torque-dependent-loss model}:
\begin{equation}
\Pthree(\vrot,\arot):=
\Ptwo(\vrot,\arot)+\ktl\taum^2
=\Jrot\vrot\arot+\ctau|\vrot|^3
+\ktl(\Jrot\arot+\ctau\vrot^2)^2,
\label{eq:p3}
\end{equation}
where $\ktl>0$ is the torque-dependent loss coefficient.

Model $\Pthree$ accounts for irreversible electromechanical losses in addition to aerodynamic dissipation and allows the acceleration-producing torque to contribute to them. Consequently, both $\Jrot$ and $\arot$ affect not only reversible kinetic-energy exchange but also irreversible dissipation, refining the attribution of the total power.

\smallskip
The preceding models contain no distinct dissipative contribution proportional to $\vrot$. However, some effects may produce an approximately speed-proportional power loss over the investigated operating domain, as occurs with an approximately constant resisting torque. Because the reversible term $\Jrot\vrot\arot$ provides the only contribution directly proportional to $\vrot$, an unmodeled loss of this form may be partially attributed to reversible power, compromising the separation between stored and dissipated power. This motivates the following \emph{four-parameter speed-proportional-loss model}:
\begin{equation}
\Pfour(\vrot,\arot)
:=\Pthree(\vrot,\arot)+\qlin|\vrot|,
\quad \qlin\geq0.
\label{eq:p4}
\end{equation}
Although numerically redundant under the assumption $\vrot\geq0$, the absolute value makes the nonnegativity of the added dissipative contribution explicit. The coefficient $\qlin$ represents a lumped speed-proportional loss over the investigated operating domain. Model $\Pfour$ can therefore represent this dissipation without assigning it to reversible kinetic-energy exchange.

\smallskip
The nested   algebraic structure can be written compactly as
\begin{equation}
\Pfour(\vrot,\arot)
=
\overbrace{
    \underbrace{
        \underbrace{\ctau|\vrot|^3}_{\Pone(\vrot)}
        \;+\;\Jrot\vrot\arot
    }_{\Ptwo(\vrot,\arot)}
    \;+\;\ktl(\Jrot\arot+\ctau\vrot^2)^2
}^{\Pthree(\vrot,\arot)}
\;+\;\qlin|\vrot|.
\label{eq:nested-power-hierarchy}
\end{equation}

Figure~\ref{fig:energetic-architecture} summarizes the energetic structure of the nested models, while Table~\ref{tab:nested-models} reports their algebraic contributions, parameters, and variables.

\section{Model-Adequacy Criteria and Methodology for Dynamic Identification and Validation}
\label{sec:method}

The models introduced in Section~\ref{sec:models} form a nested hierarchy, since each model adds one contribution to its predecessor. Because the preceding model is recovered by setting the additional coefficient to zero, the minimum fitting error cannot increase when the same data and criterion are used. The relevant question is therefore not whether a richer model fits better, but whether its additional complexity is justified by the requirements of the intended task.

In addition to predictive accuracy, model adequacy depends on faithfully attributing the predicted power. This requires distinguishing reversible kinetic-energy exchange from irreversible dissipation and, within the latter, aerodynamic from electromechanical losses. Separating reversible from dissipated power clarifies how much energy is temporarily stored in rotor motion and how much is irreversibly lost, supporting transient-energy, efficiency, and thermal assessments. Separating aerodynamic from electromechanical dissipation further indicates whether efficiency improvements should primarily target the propeller aerodynamics or the motor-drive system.

With reference to the physically motivated models introduced in Section~\ref{sec:models} and to a propeller representative of multirotor robotic platforms, this work addresses the following questions:
\begin{enumerate}[leftmargin=*]
\item Is the speed-only aerodynamic model $\Pone$ sufficiently accurate for estimating power during accelerated rotor motion?

\item If acceleration dependence is required, is the reversible kinetic-energy rate introduced by $\Ptwo$ sufficient, must the acceleration-producing torque also enter the torque-dependent electromechanical loss represented by $\Pthree$, or is the speed-proportional dissipation represented by $\Pfour$ also required?

\item How well can the dynamic models $\Ptwo$--$\Pfour$ separate reversible from dissipated power and, for $\Pthree$ and $\Pfour$, aerodynamic from electromechanical dissipation?
\end{enumerate}

To address these questions, the models are evaluated in terms of both predictive accuracy and physical power attribution. Predictive accuracy is evaluated on complete experiments excluded from parameter identification. Physical attribution is assessed by comparing the identified values of $\Jrot$ and $\ctau$ with independent aeromechanical estimates. These comparisons indicate whether the models assign the measured power consistently to inertial energy exchange, aerodynamic dissipation, and electromechanical dissipation, rather than merely predicting their sum accurately.

\subsection{Data acquisition methodology}

Assume that the electrical input power $\Pin$ delivered by the source to the motor-drive--propeller system can be determined at each time. Under a consistent sign convention, conservation of energy relates this input to the rate of stored energy and the irreversible power dissipated within the system. Identifying and evaluating the four models would ideally require sampling $\Pin$ over the entire relevant $(\vrot,\arot)$ domain. However, rotor speed and acceleration cannot be prescribed independently at arbitrary operating points, and the accessible domain is constrained by the experimental system. Assuming an internal speed controller, reference trajectories are therefore designed to sample the relevant domain through dynamically realizable experiments.

\subsubsection{Realizable operating-domain sampling}

The internal speed controller accepts a commanded rotor frequency $f_{\mathrm{cmd}}$ [Hz], related to the commanded angular speed by $\vrot_{\mathrm{cmd}}=2\pi f_{\mathrm{cmd}}$ [rad/s]. To sample positive and negative accelerations over the relevant speed range, the commanded frequency repeatedly traverses $[f_{\mathrm{low}},f_{\mathrm{high}}]$ through increasing and decreasing constant-rate ramps.

Let $\Aset=\{a_1,\ldots,a_M\}$ denote the positive nominal frequency-rate magnitudes. For each $a_j\in\Aset$, the interior of ramp $r$ is defined by
\begin{equation}
f_{\mathrm{cmd}}(t)
=f_r^0+s_ra_j(t-t_r^-),
\quad s_r\in\{-1,+1\},
\label{eq:ramp}
\end{equation}
where $t_r^-$ and $f_r^0$ are its initial time and commanded frequency, and $s_r$ determines its direction. Each $a_j$ defines one experimental condition comprising repeated ramps in both directions.

The nominal rate $s_ra_j$ defines the reference trajectory but is not a model input. Instead, the measured rotor frequency $f$ and its reconstructed derivative $\dot f$ determine
\begin{equation}
\vrot=2\pi f,
\quad
\arot=2\pi\dot f.
\label{eq:conversion}
\end{equation}
Thus, the references determine the sampled operating conditions, while model identification and validation use the realized rotor motion.

\subsubsection{Grouped identification and held-out validation}

For each $a_j\in\Aset$, let $\mathcal{E}_j$ denote the complete experiment conducted at that nominal rate. The rate set is partitioned as
\begin{equation}
\Aset=\Aid\mathbin{\sqcup}\Aval,
\end{equation}
with corresponding experiment sets
\begin{equation}
\Eid:=\{\mathcal{E}_j:a_j\in\Aid\},
\quad
\Eval:=\{\mathcal{E}_j:a_j\in\Aval\}.
\end{equation}
Every complete experiment, including all increasing and decreasing ramps, therefore belongs to either $\Eid$ or $\Eval$, and validation uses nominal rates excluded from identification.

Let $\Did$ and $\Dval$ denote the processed sample datasets obtained from $\Eid$ and $\Eval$. Each retained sample is $(\vrot,\arot,\Pin)$. Model parameters are estimated exclusively from $\Did$, while predictions on $\Dval$ quantify held-out performance on complete experiments and nominal rates excluded from identification.

\subsubsection{Signal synchronization and processing}

Let $V_{\mathrm{dc}}$ and $I_{\mathrm{dc}}$ denote the voltage and current measured at the electrical-source port, with current positive when delivered to the motor-drive--propeller system. The voltage, current, and rotor-frequency measurements are synchronized offline on a common time grid and processed using fixed noncausal zero-phase filters, identically for identification and validation.

Denoting filtered signals by a tilde, the source-side electrical power is reconstructed as
\begin{equation}
\Pin(t):=\widetilde V_{\mathrm{dc}}(t)\widetilde I_{\mathrm{dc}}(t).
\label{eq:measured-power}
\end{equation}
The filtered frequency and its derivative determine $\vrot$ and $\arot$ through \eqref{eq:conversion}. Zero-phase filtering avoids relative phase shifts among the reconstructed power, angular speed, and angular acceleration.

Transition regions are excluded because the commanded frequency is not affine when its rate changes sign. The quality criteria specified in Section~\ref{sec:signal-processing-quality} are then applied to each complete ramp interior before model fitting. If a ramp fails, its entire interior is discarded, preventing model residuals from influencing data selection.

\subsection{Parameter estimation and prediction metrics}

\subsubsection{Fitting algorithms}

For each model $P_i$, $i\in\{1,2,3,4\}$, the parameters are estimated from the identification dataset $\Did$ by minimizing the sum of squared differences between the measured source-side power $\Pin$ and the corresponding model prediction.  The parameter arrays are
\[
(\ctau), \quad
(\ctau,\Jrot), \quad
(\ctau,\Jrot,\ktl), \quad
(\ctau,\Jrot,\ktl,\qlin)
\]
for $\Pone$, $\Ptwo$, $\Pthree$, and $\Pfour$, respectively.

The models $\Pone$ and $\Ptwo$ are linear in their parameters and are fitted by scaled nonnegative linear least squares. The models $\Pthree$ and $\Pfour$ are nonlinear because $\ctau$ and $\Jrot$ also enter the squared motor-torque term; they are therefore fitted by constrained nonlinear least squares using a positive parameterization.

The nonlinear optimization is initialized independently of the external estimates of $\Jrot$ and $\ctau$ described in Section~\ref{sec:independent}. These estimates are neither included in the fitting objective nor imposed as parameter targets. The fitted parameters are subsequently held fixed when each model is evaluated on the validation dataset $\Dval$.

\subsubsection{Prediction metrics}

After identification, each model $\Pin_\ell$, $\ell\in\{1,2,3,4\}$, is evaluated on $\Dval$ with its identified parameters. Let $\Pin^{(k)}$ denote the measured source-side power at validation sample $k$, and let $\Phat_\ell^{(k)}$ denote the corresponding prediction of model $\Pin_\ell$. The prediction error is
\begin{equation}
e_\ell^{(k)}
:=
\Phat_\ell^{(k)}-\Pin^{(k)}.
\label{eq:prediction-error}
\end{equation}
For the $N$ samples in $\Dval$, the global prediction metrics of model $\Pin_\ell$ are
\begin{equation}
\operatorname{RMSE}_\ell
:=
\sqrt{\frac{1}{N}\sum_{k=1}^N
\left(e_\ell^{(k)}\right)^2},
\quad
\operatorname{MAE}_\ell
:=
\frac{1}{N}\sum_{k=1}^N
\left|e_\ell^{(k)}\right|,
\label{eq:rmse-mae}
\end{equation}
and
\begin{equation}
R_\ell^2
:=
1-
\left(\sum_{k=1}^N\left(e_\ell^{(k)}\right)^2\right)
\;\big/\;
\left(\sum_{k=1}^N\left(\Pin^{(k)}-\overline{\Pin}\right)^2\right).
\label{eq:r2}
\end{equation}
where $\overline{\Pin}:=N^{-1}\sum_{k=1}^N\Pin^{(k)}$.

Because relative errors become ill-conditioned when the measured power approaches zero, relative predictive accuracy is evaluated only for samples satisfying $\Pin^{(k)}\geq\Pmin$, where $\Pmin>0$ is a prescribed threshold. Let $N_{\min}$ denote the number of such samples. The corresponding mean absolute percentage error is
\begin{equation}
\operatorname{MAPE}_{\ell,\Pin\geq\Pmin}
:=
\frac{100}{N_{\min}}
\sum_{k:\Pin^{(k)}\geq\Pmin}
\left|\Phat_\ell^{(k)}-\Pin^{(k)}\right|
\big/
\Pin^{(k)}.
\label{eq:mape}
\end{equation}
This metric reports the average absolute relative prediction error over the selected positive-power region; it does not bound individual prediction errors.

\subsubsection{Experiment-level statistical and local assessment}

The sampling uncertainty associated with each predictive improvement is assessed by resampling the complete experiments in $\Eval$ with replacement and recomputing the validation metrics on the resulting samples. Resampling complete experiments rather than individual samples preserves the dependence among measurements collected along the same trajectory.

To determine where each model extension improves prediction, the sampled $(\vrot,\arot)$ domain of $\Dval$ is partitioned into cells. For two consecutive models, with model $B$ denoting the richer model and model $A$ its predecessor, the local relative RMSE improvement is
\begin{equation}
\mathcal I_{B/A}
:=
1-\operatorname{RMSE}_{B}/
\operatorname{RMSE}_{A},
\label{eq:local-improvement}
\end{equation}
where both RMSE values are computed from the validation samples within the same cell. The metric is evaluated only in sufficiently populated cells. A positive value of $\mathcal I_{B/A}$ indicates that model $B$ has a lower local RMSE than model $A$.

\subsection{Independent aeromechanical validation}
\label{sec:independent}

Held-out prediction assesses how accurately each identified model estimates the measured electrical power, but it does not establish whether that power is correctly decomposed into reversible and dissipative contributions. In particular, different values of $\Jrot$, $\ctau$, and the electrical-loss coefficients may compensate for one another while producing similar total-power predictions. We therefore determine $\Jrot$ and $\ctau$ independently of the electrical-power measurements and use them as independent external references for evaluating the power attribution provided by each model.

\subsubsection{Independent inertia determination}

An independent estimate $\Jrot^{\mathrm{ind}}$ is obtained from the inertias of the rotating components. For a segmented representation of the propeller, with segment mass $m_s$ concentrated at radial distance $r_s$ from the rotation axis,
\begin{equation}
J_{\mathrm{prop}}
:=\sum_{s=1}^{S}m_sr_s^2,
\quad
\Jrot^{\mathrm{ind}}
:=J_{\mathrm{motor}}+J_{\mathrm{prop}}+J_{\mathrm{other}},
\label{eq:independent-J}
\end{equation}
where $J_{\mathrm{motor}}$ and $J_{\mathrm{other}}$ denote the inertias of the motor rotor and any other rotating components, respectively. This estimate is obtained independently of the electrical-power data used to identify $\Pone$--$\Pfour$.

\subsubsection{Independent aerodynamic-drag determination}

An independent torque measurement is used to identify the quasi-steady relation between aerodynamic resisting torque and rotor frequency,
\begin{equation}
\taud(f)=c_{\tau,f}^{\mathrm{ind}}f^2.
\label{eq:independent-drag-frequency}
\end{equation}
Since $\vrot=2\pi f$, comparison with \eqref{eq:drag-torque} gives
\begin{equation}
\ctau^{\mathrm{ind}}
:=c_{\tau,f}^{\mathrm{ind}}/ (2\pi)^2.
\label{eq:ctau-conversion}
\end{equation}
This estimate is likewise obtained independently of the electrical-power data and provides an external reference for the aerodynamic contribution identified by each model.

\subsubsection{Joint validation criterion}

For an aeromechanical parameter $\theta\in\{\Jrot,\ctau\}$, let $\widehat{\theta}$ denote its estimate obtained from electrical-power identification and $\theta^{\mathrm{ind}}$ its independent estimate. Their signed relative discrepancy is
\begin{equation}
\delta_\theta
:=100
\,(\widehat{\theta}-\theta^{\mathrm{ind}})/
(\theta^{\mathrm{ind}}).
\label{eq:parameter-discrepancy}
\end{equation}
Accordingly, $\delta_\theta>0$ indicates overestimation and $\delta_\theta<0$ indicates underestimation relative to the independent value.

\medskip
The held-out prediction metrics and the parameter discrepancies assess complementary properties. RMSE, MAE, $R^2$, and MAPE quantify the accuracy of the predicted total electrical power, whereas $\delta_{\Jrot}$ and $\delta_{\ctau}$ assess whether the model attributes the inertial and aerodynamic contributions consistently with their independent estimates. Accurate total-power prediction does not imply physically accurate attribution, and neither validation criterion replaces the other.

\section{Experimental Realization and Dataset}
\label{sec:implementation}

This section describes the experimental realization of the methodology in Section~\ref{sec:method} for the particular hardware setup used. It specifies the propulsion unit, instrumentation, trajectory family, signal-processing settings, and independent aeromechanical references used to obtain the reported results.

\subsection{Propulsion unit and data acquisition}

\subsubsection{Hardware and instrumentation}

The tested propulsion unit comprises an Xnova Lightning 2208 V2 1500KV motor, a Graupner 3D-Prop 8x4.5 propeller, and an Aikon AKC 70A 8S electronic speed controller. The principal hardware characteristics are reported in Table~\ref{tab:hardware}.

\begin{table*}[t]
\centering
\caption{Experimental hardware and instrumentation.}
\label{tab:hardware}
\begin{tabular}{
    p{0.18\textwidth}
    p{0.27\textwidth}
    p{0.47\textwidth}}
\toprule
\textbf{Component}
&
\textbf{Device}
&
\textbf{Relevant characteristic}
\\
\midrule
Motor
&
Xnova Lightning 2208 V2 1500KV
&
1500~rpm/V; listed resistance 0.078~$\Omega$; mass 36.5~g
\\
Propeller
&
Graupner 3D-Prop 8x4.5
&
8-in diameter; 4.5-in pitch; measured mass 7.81~g
\\
Electronic speed controller
&
Aikon AKC 70A 8S 8-in-1
&
BLHeli\_32; 70-A continuous rating
\\
Current sensing
&
Mateksys Hall Current Sensor 150A
&
20~mV/A sensitivity; 5\% stated current-output error
\\
Acquisition and logging
&
D1 Mini, Chimera board, and Raspberry Pi
&
Timestamped current, voltage, and rotor-frequency acquisition
\\
\bottomrule
\end{tabular}
\end{table*}

\ifextendedversion
Figure~\ref{fig:hardware} shows the propulsion and acquisition components.

\begin{figure}[t]
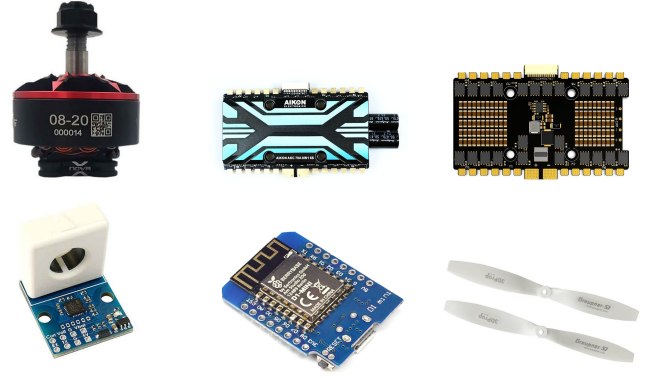

\centering
\begin{minipage}[t]{0.30\columnwidth}
    \centering
    \safegraphic[width=\linewidth]{\figHardwareA}
\end{minipage}\hfill
\begin{minipage}[t]{0.30\columnwidth}
    \centering
    \safegraphic[width=\linewidth]{\figHardwareB}
\end{minipage}\hfill
\begin{minipage}[t]{0.30\columnwidth}
    \centering
    \safegraphic[width=\linewidth]{\figHardwareC}
\end{minipage}

\vspace{0.3em}

\begin{minipage}[t]{0.30\columnwidth}
    \centering
    \safegraphic[width=\linewidth]{\figHardwareD}
\end{minipage}\hfill
\begin{minipage}[t]{0.30\columnwidth}
    \centering
    \safegraphic[width=\linewidth]{\figHardwareE}
\end{minipage}\hfill
\begin{minipage}[t]{0.30\columnwidth}
    \centering
    \safegraphic[width=\linewidth]{\figHardwareF}
\end{minipage}

\caption{Propulsion and measurement components used in the experiments: motor-propeller unit, electronic speed controller, current sensor and acquisition electronics, voltage acquisition, and logging computer.}
\label{fig:hardware}
\end{figure}
\fi

The source-side DC current was measured using a Mateksys 150A Hall current sensor and acquired through a D1 Mini. The battery voltage and rotor frequency were obtained from the low-level software log. The voltage and current acquisition systems were connected to the same Raspberry Pi, and their logs were synchronized offline using Unix timestamps.

\subsubsection{Trajectory family and dataset}

The commanded rotor-frequency interval was $[35,220]$~Hz. The lower bound avoids the reduced accuracy of the sensorless rotor-speed estimate near its minimum reliable frequency, while the upper bound avoids current saturation. Identification and validation rates were interleaved across this interval, with the extreme rates included in identification, to evaluate prediction at unseen rates within the sampled dynamic domain. The nominal frequency-rate magnitudes assigned to identification were
\[
\begin{aligned}
\Aid
={}&\{2,10,50,100,150,200,250,\\
   &\phantom{\{}350,450,700,900,1250,2000\}\ \mathrm{Hz/s},
\end{aligned}
\]
whereas those assigned to held-out validation were
\[
\begin{aligned}
\Aval
={}&\{25,75,125,175,225,300,\\
   &\phantom{\{}400,500,600,800,1000,1500\}\ \mathrm{Hz/s}.
\end{aligned}
\]
Thus, thirteen complete experiments were used for identification and thirteen distinct experiments for validation.

After excluding the 25-Hz reversal margins and applying quality control, the 26 experiments yielded 278,627 retained samples. The measured rotor frequency ranged from 60.28 to 195.05~Hz, the reconstructed frequency rate from $-1489.83$ to 2093.45~Hz/s, and the measured source-side power from $-16.60$ to 841.76~W. The validation dataset comprised 140,508 samples.

\subsubsection{Signal processing and quality control}
\label{sec:signal-processing-quality}

The rotor-frequency, voltage, and current signals were processed offline using third-order noncausal zero-phase Savitzky--Golay filters. The rotor-frequency filter used a window of 101 samples, while the voltage and current filters used windows of 31 samples. A 25-Hz margin from each commanded reversal excluded the transition regions in which the affine-ramp model \eqref{eq:ramp} does not apply.

Quality control was performed on complete ramp interiors before fitting any power model. A ramp interior was rejected when the discrepancy between the measured and filtered rotor frequencies exceeded 10~Hz or when the realized-rate quality criterion was not satisfied.

The internal speed-controller gains remained fixed throughout the complete identification and validation campaign.

\ifextendedversion
Figure~\ref{fig:ramp-quality} illustrates the resulting exclusion, acceptance, and rejection of complete ramp interiors.

\begin{figure}[t!]
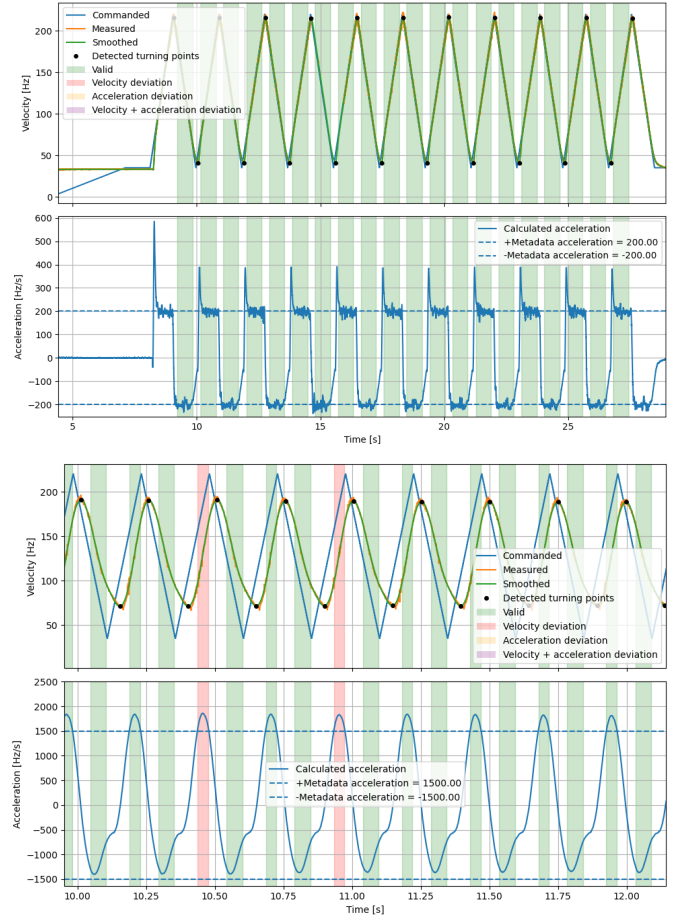

\centering
\begin{minipage}[t]{0.49\textwidth}
    \centering
    \safegraphic[width=\linewidth]{\figCleaning}
\end{minipage}\\
\begin{minipage}[t]{0.49\textwidth}
    \centering
    \safegraphic[width=\linewidth]{\figRejected}
\end{minipage}
\caption{Ramp-level quality control. Transition regions around commanded frequency reversals are excluded, and complete ramp interiors are retained or rejected according to measured-motion and processing-consistency criteria established before fitting any power model.}
\label{fig:ramp-quality}
\end{figure}
\fi

\subsection{Independent references and validation settings}

\subsubsection{Rotational-inertia reference}

\begin{figure}[t]
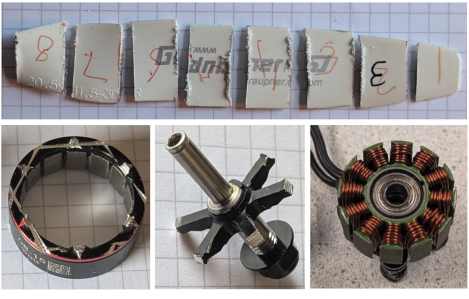

\centering
\safegraphic[width=0.70\linewidth]
{\figInertiaSeparated}

\caption{Independent determination of the rotational inertia:
(a) separation of one propeller blade from the hub;
(b) division of the blade into eight segments for radial
mass-distribution measurement; and
(c) disassembly of the motor to characterize the outer rotor and shaft.}
\label{fig:inertia-measurement}
\end{figure}

The independent inertia was determined by destructively disassembling the propeller and motor rotor. One propeller blade was separated from the hub and divided into eight segments to account for its nonuniform radial mass distribution. The individual segment masses and lengths were measured and corrected for the approximately 2\% mass and length lost during cutting. The motor rotor was separated from the stator and shaft; the outer rotor was approximated as an annular cylinder and the shaft as a solid cylinder. Figure~\ref{fig:inertia-measurement} summarizes this procedure.

For each corrected blade segment, the inertia was computed from its mass and inner and outer radial positions according to \eqref{eq:independent-J}. The two blades and hub yielded
\[
J_{\mathrm{prop}}
=
1.8820\times10^{-5}\ \mathrm{kg\,m^2}.
\]
The motor rotor and shaft contributed
\[
J_{\mathrm{motor}}
=
1.921\times10^{-6}\ \mathrm{kg\,m^2},
\quad
J_{\mathrm{shaft}}
=
1.04\times10^{-8}\ \mathrm{kg\,m^2},
\]
respectively. Their sum gave the independent total-inertia estimate
\begin{equation}
\Jrot^{\mathrm{ind}}
=
2.075\times10^{-5}\ \mathrm{kg\,m^2}.
\label{eq:J-value}
\end{equation}

\subsubsection{Aerodynamic-drag reference and validation threshold}

The independent force-torque measurement yielded
\[
c_{\tau,f}^{\mathrm{ind}}
=
2.8220\times10^{-6}\ \mathrm{N\,m\,Hz^{-2}}.
\]
Using \eqref{eq:ctau-conversion}, the corresponding angular-speed coefficient is
\begin{equation}
\ctau^{\mathrm{ind}}
=
7.14821\times10^{-8}\
{\mathrm{N\,m}}/{(\mathrm{rad/s})^2}.
\label{eq:ctau-value}
\end{equation}

For the relative prediction metric in \eqref{eq:mape}, the threshold was set to $\Pmin=20$~W. This condition retained 109,282 of the 140,508 validation samples.

\section{Experimental Results}
\label{sec:results}

This section evaluates the nested models according to the two complementary criteria introduced in Section~\ref{sec:method}: held-out predictive accuracy and recovery of the independently determined aeromechanical parameters.

\subsection{Held-out prediction along the nested hierarchy}

Table~\ref{tab:nested-results} reports the held-out prediction metrics. The transition from $\Pone$ to $\Ptwo$ reduces RMSE by 58.9\%; the median experiment-level improvement is 58.9\%, with a 95\% bootstrap interval of 55.7--61.8\%. Thus, speed alone is insufficient and the reversible kinetic-energy rate is necessary. The transition to $\Pthree$ reduces RMSE by a further 53.4\%, confirming that reversible power alone is insufficient and that the acceleration-producing torque must also contribute to electromechanical dissipation. Model $\Pfour$ provides the lowest errors, with a smaller final improvement through lumped speed-proportional dissipation.

\begin{table}[t]
\centering
\caption{Held-out prediction metrics along the nested model hierarchy.}
\label{tab:nested-results}
\begin{tabular}{lrrrr}
\toprule
\textbf{Model}
&
\textbf{RMSE}
&
\textbf{MAE}
&
$\boldsymbol{R^2}$
&
\textbf{MAPE}
\\
&
[W]
&
[W]
&
&
[\%]
\\
\midrule
$\Pone$   & 103.320 & 61.785 & 0.18044 & 53.945 \\
$\Ptwo$   &  42.457 & 26.602 & 0.86161 & 16.579 \\
$\Pthree$ &  19.799 & 12.465 & 0.96991 & 14.687 \\
$\Pfour$  &  17.388 &  9.651 & 0.97679 &  8.739 \\
\bottomrule
\end{tabular}
\end{table}

\begin{figure*}[t]
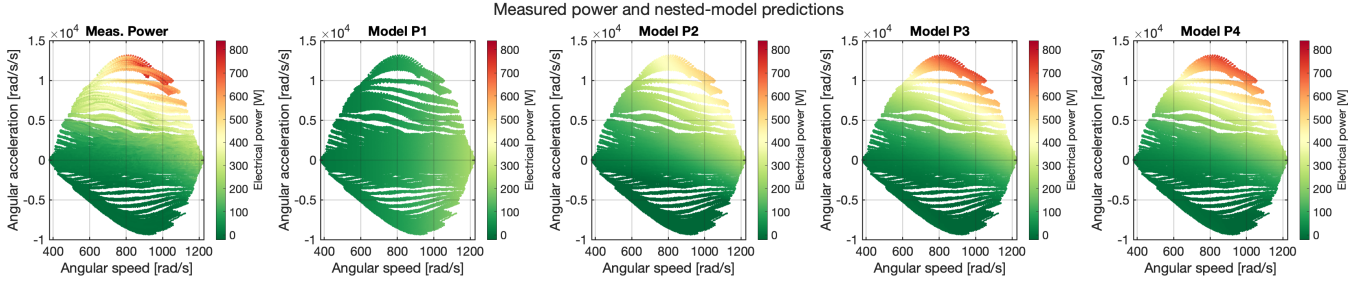

\centering
\safegraphic[width=0.99\textwidth]{\figNestedPower}
\caption{Measured source-side electrical power and held-out predictions of the nested models over the sampled angular-speed and acceleration domain. The common color scale shows the progressive recovery of acceleration-dependent power, with $\Pfour$ providing the closest overall prediction among the nested models.}
\label{fig:nested-power}
\end{figure*}

Figure~\ref{fig:nested-power} compares the measured source-side electrical power with the predictions of the four models over the sampled $(\vrot,\arot)$ domain. The common color scale shows the progressive recovery of the acceleration-dependent power variation.

\begin{figure*}[t]
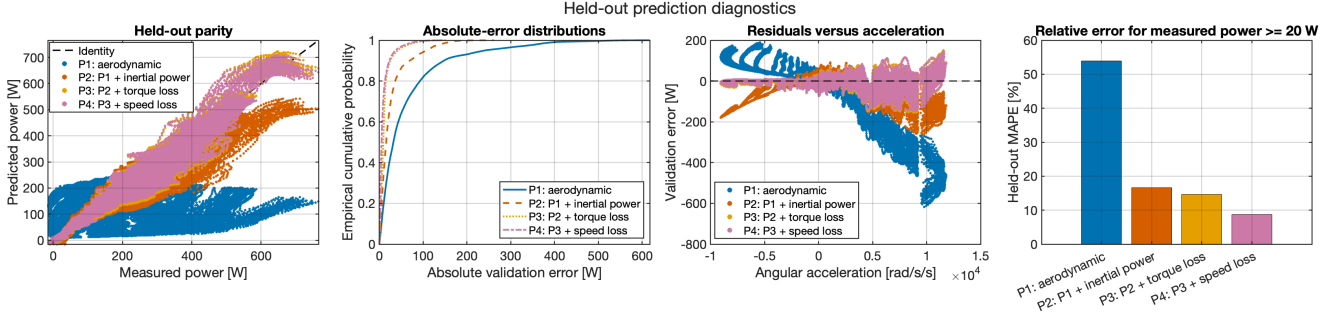

\centering
\safegraphic[width=0.99\textwidth]{\figNestedDiagnostics}
\caption{Held-out prediction diagnostics for the nested models: prediction parity, cumulative absolute-error distributions, residual dependence on angular acceleration, and MAPE for measured source-side power satisfying $\Pin\geq20$~W. The principal improvements occur through $\Pthree$, while $\Pfour$ provides a smaller refinement and therefore overlaps closely with $\Pthree$ in some panels.}
\label{fig:nested-diagnostics}
\end{figure*}

\begin{figure*}[t]
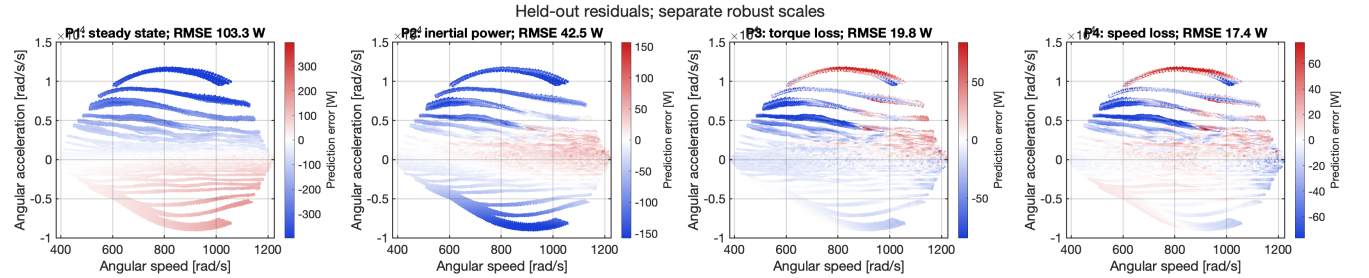

\centering
\safegraphic[width=0.98\textwidth]{\figNestedResiduals}
\caption{Held-out prediction residuals, defined as predicted minus measured source-side power. The strong acceleration-organized structure observed for $\Pone$ is progressively reduced by the dynamic models, with $\Pthree$ providing the principal correction and $\Pfour$ a smaller additional refinement. Each panel uses a separate symmetric color scale based on the 99th percentile of the corresponding absolute residuals; color magnitudes must therefore be interpreted using the panel-specific colorbar.}
\label{fig:nested-residuals}
\end{figure*}

Figure~\ref{fig:nested-diagnostics} reports complementary held-out diagnostics, including prediction parity, cumulative absolute-error distributions, residual dependence on angular acceleration, and MAPE for $\Pin\geq20$~W. Figure~\ref{fig:nested-residuals} further shows that the large acceleration-organized residuals of $\Pone$ are progressively reduced by the dynamic models. The main improvements occur from $\Pone$ to $\Ptwo$ and from $\Ptwo$ to $\Pthree$, whereas the predictions of $\Pthree$ and $\Pfour$ largely overlap. The residual dispersion is smaller over part of the negative-acceleration region than at corresponding positive accelerations. This asymmetry is already present in the measured-power distribution: negative acceleration is associated with a comparatively narrow power range, whereas positive acceleration produces a broader and more strongly acceleration-dependent response. The models consequently reproduce the former region more uniformly, while the latter retains greater residual variability.

\subsection{Incremental improvement over the operating domain}

\begin{figure*}[t]
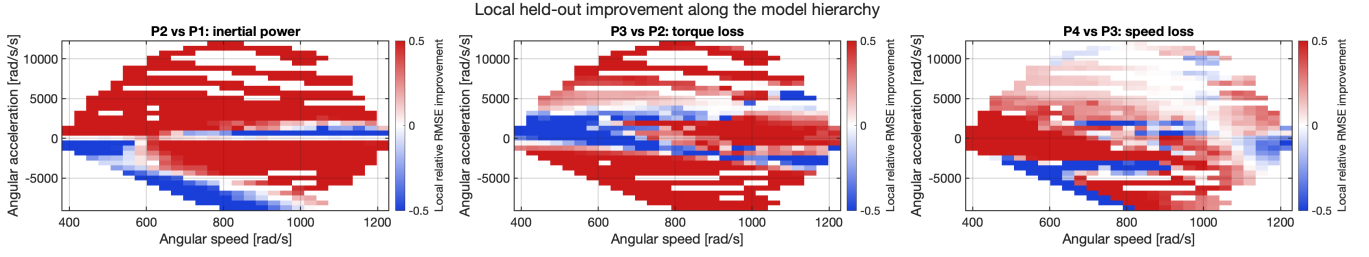

\centering
\safegraphic[width=0.99\textwidth]{\figNestedImprovements}
\caption{Local held-out RMSE improvement across the nested hierarchy. The panels show the effects of introducing reversible kinetic-energy rate, torque-dependent electromechanical dissipation, and lumped speed-proportional dissipation. Positive values indicate a lower local RMSE for the richer model. Uncolored cells contain insufficient validation samples.}
\label{fig:nested-improvements}
\end{figure*}

The global metrics establish the overall benefit of each additional contribution, while Fig.~\ref{fig:nested-improvements} shows where this benefit occurs within the sampled $(\vrot,\arot)$ domain. The local improvement is evaluated according to \eqref{eq:local-improvement} for each pair of consecutive models.

The transition from $\Pone$ to $\Ptwo$ produces the dominant correction across the accelerated operating domain. The transition from $\Ptwo$ to $\Pthree$ provides a second broad improvement, particularly where the motor torque is large. The transition from $\Pthree$ to $\Pfour$ provides a smaller and more localized refinement. Positive values indicate a lower local RMSE for the richer model, while uncolored cells contain insufficient validation samples.

\subsection{Identified coefficients and aeromechanical parameter recovery}

Held-out prediction assesses the reconstructed total electrical power but does not establish whether the power contributions are physically attributed correctly. Table~\ref{tab:parameter-recovery} therefore reports the complete parameter array identified for each nested model and compares the power-identified values of $\Jrot$ and $\ctau$ with the independent point estimates reported in Section~\ref{sec:implementation}.

Model $\Pone$ cannot estimate $\Jrot$, while $\Ptwo$ substantially improves prediction but does not recover either aeromechanical reference closely. Model $\Pthree$ closely recovers $\ctau$ but retains a substantial inertia discrepancy, whereas $\Pfour$ provides the closest joint recovery of the two independent point estimates. As shown in Table~\ref{tab:parameter-recovery}, the complete parameter array is jointly re-estimated at every level; $\Pfour$ is therefore not obtained by retaining the $\Pthree$ coefficients and fitting only the added term.

\begin{table}[t]

\centering
\caption{Identified coefficients and agreement with the independent
aeromechanical estimates. Dashes: coefficients absent from a model.}
\label{tab:parameter-recovery}
\renewcommand{\arraystretch}{1.12}
\setlength{\tabcolsep}{3.2pt}
\begin{tabular}{lrrrrrr}
\toprule
Model
&
$\widehat{\Jrot}$
&
$\widehat{\ctau}$
&
$\widehat{\ktl}$
&
$\widehat{\qlin}$
&
$\delta_{\Jrot}$
&
$\delta_{\ctau}$
\\
&
$[10^{-5}]$
&
$[10^{-8}]$
&
$[10^{3}]$
&
$[10^{-2}]$
&
$[\%]$
&
$[\%]$
\\
\midrule
$\Pone$
&
--
&
14.01
&
--
&
--
&
--
&
$+96.00$
\\
$\Ptwo$
&
3.570
&
13.99
&
--
&
--
&
$+72.03$
&
$+95.73$
\\
$\Pthree$
&
1.584
&
7.396
&
7.820
&
--
&
$-23.68$
&
$+3.47$
\\
$\Pfour$
&
1.902
&
7.155
&
5.229
&
2.176
&
$-8.36$
&
$+0.10$
\\
\midrule
Independent reference
&
2.075
&
7.148
&
--
&
--
&
--
&
--
\\
\bottomrule
\end{tabular}

\vspace{1mm}
\footnotesize
Units: $\widehat{\Jrot}$: $\mathrm{kg\,m^2}$,
$\widehat{\ctau}$: $\mathrm{N\,m}/(\mathrm{rad/s})^2$,
$\widehat{\ktl}$: $\mathrm{W}/(\mathrm{N\,m})^2$, and
$\widehat{\qlin}$: $\mathrm{W}/(\mathrm{rad/s})$.

\end{table}

\begin{table*}[t]
\centering

\end{table*}

\ifextendedversion
Figure~\ref{fig:nested-summary} presents the two validation criteria separately. The predictive metrics improve along the complete hierarchy, but the parameter discrepancies reveal how the additional dissipative contributions change the physical attribution of the measured power. The transition to $\Pthree$ provides the principal improvement in dynamic prediction and aerodynamic-drag recovery, whereas $\Pfour$ provides the closest joint recovery of $\Jrot$ and $\ctau$.

\begin{figure*}[t]
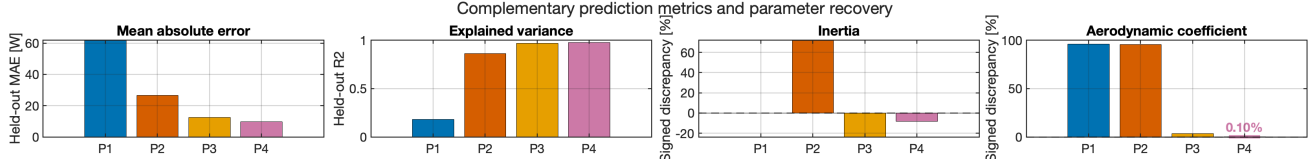

\centering
\safegraphic[width=0.99\textwidth]{\figNestedSummary}
\caption{Held-out prediction and aeromechanical parameter recovery along the nested hierarchy. Prediction metrics quantify total-power accuracy, whereas the signed parameter discrepancies quantify agreement with the independent inertia and aerodynamic-drag estimates.}
\label{fig:nested-summary}
\end{figure*}
\fi

\subsection{Decomposition of reversible and dissipative power}

\begin{figure*}[t]
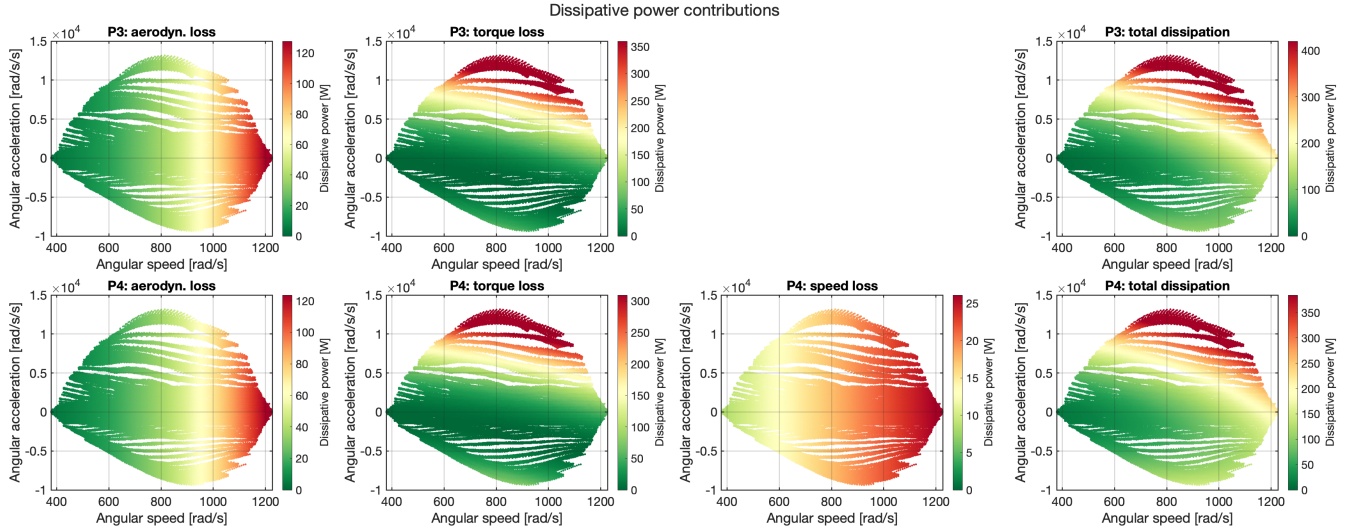

\centering
\safegraphic[width=0.98\textwidth]{\figNestedLosses}
\caption{Power decomposition obtained with $\Pthree$ and $\Pfour$. Both models separate reversible kinetic-energy rate from aerodynamic and torque-dependent electromechanical dissipation, while $\Pfour$ additionally represents lumped speed-proportional dissipation. Component-specific robust color scales are used to reveal contributions of different magnitudes; quantitative color comparisons must therefore use the corresponding colorbars.}
\label{fig:nested-losses}
\end{figure*}

Figure~\ref{fig:nested-losses} compares the power decompositions obtained with $\Pthree$ and $\Pfour$. Both models distinguish the reversible kinetic-energy rate $\Jrot\vrot\arot$ from irreversible power dissipation and separate aerodynamic dissipation from torque-dependent electromechanical dissipation. Model $\Pfour$ additionally assigns speed-proportional dissipation to the term $\qlin|\vrot|$, rather than allowing this contribution to be partially represented through the reversible kinetic-energy-rate term.

As shown by the component-specific colorbar scales in Fig.~\ref{fig:nested-losses}, the speed-proportional contribution has a markedly smaller magnitude than the principal aerodynamic and torque-dependent contributions over the investigated domain.   Nevertheless, its inclusion changes the identified values of both $\Jrot$ and $\ctau$ and brings them closer to their independent point estimates. This result supports the role of $\Pfour$ in separating reversible kinetic-energy rate from unmodeled speed-proportional dissipation.

\section{Discussion and Conclusions}
\label{sec:discussion}

The experiments answer the three questions posed in Section~\ref{sec:method}. Model $\Pone$ is inadequate during accelerated rotor motion, and the substantial improvement obtained with $\Ptwo$ confirms that the reversible kinetic-energy rate cannot be neglected. The remaining acceleration-organized errors show that this contribution alone is insufficient. Model $\Pthree$ supplies the missing torque-dependent electromechanical dissipation, providing a second substantial predictive improvement and closely recovering the aerodynamic resisting-torque coefficient, although an inertia discrepancy remains. Model $\Pfour$ adds a distinct speed-proportional dissipative channel, yielding a smaller predictive refinement but the closest joint recovery of the independent aeromechanical point estimates. It therefore provides the best combined prediction and power attribution among the considered models.

\subsection{Scope and limitations}

The experiments cover positive rotor frequencies from 60.28 to 195.05~Hz and reconstructed frequency rates from $-1489.83$ to 2093.45~Hz/s. The conclusions therefore concern interpolation within this experimental domain and the particular motor-drive--propeller unit considered. The reported MAPE values apply only to measured source-side powers satisfying $\Pin\geq20$~W.

The independent values of $\Jrot$ and $\ctau$ are point estimates obtained from segmented inertia determination and force--torque measurements, respectively. Consequently, the reported parameter discrepancies quantify agreement with these reference values but do not establish statistical compatibility. The interpretation of parameter recovery is therefore limited to the relative proximity of the identified and independently determined point estimates.

The remaining acceleration-organized residuals indicate that the four considered contributions do not describe all power variations. Possible unmodeled effects include braking and DC-bus energy return, temperature dependence, electronic-speed-controller switching and conduction losses, iron losses, voltage dependence, and residual synchronization or differentiation errors. Moreover, $\ktl$ and $\qlin$ are lumped coefficients over the investigated operating domain and should not be interpreted as isolated microscopic loss parameters.

\subsection{Implications and conclusion}
Speed-only models may remain adequate near steady operation but can become strongly inaccurate during rapid rotor-speed modulation. Models $\Pthree$ and $\Pfour$ remain compact enough for energy-aware trajectory generation, control allocation, and propulsion-system analysis while representing both reversible kinetic-energy exchange and acceleration-dependent dissipation. This distinction clarifies how much energy is stored temporarily in rotor motion and how much is lost through aerodynamic and electromechanical dissipation, supporting transient-energy estimation, efficiency assessment, and interpretation of identified propulsion parameters.

This work developed and experimentally evaluated a nested sequence of propulsion-power models progressing from aerodynamic power dissipation to reversible kinetic-energy exchange, torque-dependent electromechanical dissipation, and lumped speed-proportional dissipation. Held-out prediction establishes that the kinetic-energy rate is necessary but insufficient and that torque-dependent dissipation provides the principal additional improvement. The speed-proportional contribution further improves prediction and provides the closest joint recovery of the independently determined inertia and aerodynamic coefficient. Thus, held-out prediction and independent aeromechanical references assess complementary properties and together provide a physically grounded basis for selecting dynamic propulsion-power models.

\section*{Acknowledgment}
The authors thank the technical staff supporting the experiments: Quentin Sablé and Artur Gąsienica.

\bibliographystyle{IEEEtran}
\bibliography{refs}

\begin{thebibliography}{10}
\providecommand{\url}[1]{#1}
\csname url@samestyle\endcsname
\providecommand{\newblock}{\relax}
\providecommand{\bibinfo}[2]{#2}
\providecommand{\BIBentrySTDinterwordspacing}{\spaceskip=0pt\relax}
\providecommand{\BIBentryALTinterwordstretchfactor}{4}
\providecommand{\BIBentryALTinterwordspacing}{\spaceskip=\fontdimen2\font plus
\BIBentryALTinterwordstretchfactor\fontdimen3\font minus
  \fontdimen4\font\relax}
\providecommand{\BIBforeignlanguage}[2]{{%
\expandafter\ifx\csname l@#1\endcsname\relax
\typeout{** WARNING: IEEEtran.bst: No hyphenation pattern has been}%
\typeout{** loaded for the language `#1'. Using the pattern for}%
\typeout{** the default language instead.}%
\else
\language=\csname l@#1\endcsname
\fi
#2}}
\providecommand{\BIBdecl}{\relax}
\BIBdecl

\bibitem{Liu2017Power}
Z.~Liu, R.~Sengupta, and A.~Kurzhanskiy, ``A power consumption model for
  multi-rotor small unmanned aircraft systems,'' in \emph{2017 International
  Conference on Unmanned Aircraft Systems (ICUAS)}, 2017, pp. 310--315.

\bibitem{Bauersfeld2022Range}
L.~Bauersfeld and D.~Scaramuzza, ``Range, endurance, and optimal speed
  estimates for multicopters,'' \emph{IEEE Robotics and Automation Letters},
  vol.~7, no.~2, pp. 2953--2960, 2022.

\bibitem{Gong2023Power}
H.~Gong, B.~Huang, B.~Jia, and H.~Dai, ``Modeling power consumptions for
  multirotor {UAVs},'' \emph{IEEE Transactions on Aerospace and Electronic
  Systems}, vol.~59, no.~6, pp. 7409--7422, 2023.

\bibitem{AguilarLopez2022Energy}
J.~M. Aguilar-L{\'o}pez, R.~A. Garc{\'i}a, C.~Bordons, and E.~F. Camacho,
  ``Development of the energy consumption model of a quadrotor using voltage
  data from experimental flights,'' in \emph{2022 17th IEEE International
  Conference on Control \& Automation (ICCA)}, 2022, pp. 432--437.

\bibitem{Dai2019Efficiency}
X.~Dai, Q.~Quan, J.~Ren, and K.-Y. Cai, ``Efficiency optimization and component
  selection for propulsion systems of electric multicopters,'' \emph{IEEE
  Transactions on Industrial Electronics}, vol.~66, no.~10, pp. 7800--7809,
  2019.

\bibitem{Kotarski2023HeavyLift}
D.~Kotarski, P.~Piljek, and J.~Kasa{\v{c}}, ``Heavy-lift multirotor unmanned
  aerial vehicles characterization and sizing,'' \emph{IEEE Access}, vol.~11,
  pp. 95\,385--95\,405, 2023.

\bibitem{Joo2025Hybrid}
S.~Joo, H.~Kim, and H.~Son, ``Multiobjective design optimization of
  hybrid-powered multirotors for enhanced endurance and payload capacity,''
  \emph{IEEE/ASME Transactions on Mechatronics}, vol.~30, no.~3, pp.
  1959--1970, 2025.

\bibitem{Coates2019Propulsion}
E.~M. Coates, A.~Wenz, K.~Gryte, and T.~A. Johansen, ``Propulsion system
  modeling for small fixed-wing {UAVs},'' in \emph{2019 International
  Conference on Unmanned Aircraft Systems (ICUAS)}, 2019, pp. 748--757.

\bibitem{Montoya2021Dynamic}
M.~M. Rivera and L.~B. Gutierrez, ``Dynamic model of an aircraft propulsion
  system with electric motor and propeller,'' in \emph{2021 IEEE 5th Colombian
  Conference on Automatic Control (CCAC)}, 2021, pp. 157--162.

\bibitem{Alkomy2023Trajectories}
H.~Alkomy and J.~Shan, ``Investigating the effects of polynomial trajectories
  on energy consumption of quadrotors,'' \emph{IEEE/ASME Transactions on
  Mechatronics}, vol.~28, no.~3, pp. 1593--1604, 2023.

\bibitem{Nascimento2019Survey}
T.~P. Nascimento and M.~Saska, ``Position and attitude control of multi-rotor
  aerial vehicles: A survey,'' \emph{Annual Reviews in Control}, vol.~48, pp.
  129--146, 2019.

\bibitem{Baca2021MRS}
T.~Baca, M.~Petrlik, M.~Vrba, V.~Spurny, R.~Penicka, D.~Hert, and M.~Saska,
  ``The {MRS UAV} system: Pushing the frontiers of reproducible research,
  real-world deployment, and education with autonomous unmanned aerial
  vehicles,'' \emph{Journal of Intelligent \& Robotic Systems}, vol. 102,
  no.~1, p.~26, 2021.

\bibitem{Ruggiero2018Review}
F.~Ruggiero, V.~Lippiello, and A.~Ollero, ``Aerial manipulation: A literature
  review,'' \emph{IEEE Robotics and Automation Letters}, vol.~3, no.~3, pp.
  1957--1964, 2018.

\bibitem{Ollero2022Past}
A.~Ollero, M.~Tognon, A.~Suarez, D.~Lee, and A.~Franchi, ``Past, present, and
  future of aerial robotic manipulators,'' \emph{IEEE Transactions on
  Robotics}, vol.~38, no.~1, pp. 626--645, 2022.

\bibitem{Ryll2017Interaction}
M.~Ryll, G.~Muscio, F.~Pierri, E.~Cataldi, G.~Antonelli, F.~Caccavale, and
  A.~Franchi, ``{6D} physical interaction with a fully actuated aerial robot,''
  in \emph{2017 IEEE International Conference on Robotics and Automation
  (ICRA)}, 2017, pp. 5190--5195.

\bibitem{Tognon2019Redundant}
M.~Tognon, H.~A.~T. Chavez, E.~Gasparin, Q.~Sabl{\'e}, D.~Bicego, A.~Mallet,
  M.~Lany, G.~Santi, B.~Revaz, J.~Cort{\'e}s, and A.~Franchi, ``A
  truly-redundant aerial manipulator system with application to push-and-slide
  inspection in industrial plants,'' \emph{IEEE Robotics and Automation
  Letters}, vol.~4, no.~2, pp. 1846--1851, 2019.

\bibitem{Dyer2019Energy}
E.~Dyer, S.~Sirouspour, and M.~Jafarinasab, ``Energy optimal control allocation
  in a redundantly actuated omnidirectional {UAV},'' in \emph{2019
  International Conference on Robotics and Automation (ICRA)}, 2019, pp.
  5316--5322.

\bibitem{Taylor2024Reconfigurable}
J.~Taylor, N.~Imanberdiyev, M.~Y.~M. Chuah, W.-Y. Yau, G.~Sartoretti, and
  E.~Camci, ``Reconfigurable multi-rotor for high-precision physical
  interaction,'' in \emph{2024 IEEE/RSJ International Conference on Intelligent
  Robots and Systems (IROS)}, 2024, pp. 8069--8074.

\bibitem{Veenstra2026Interaction}
R.~Veenstra, A.~Ali, C.~Gabellieri, and A.~Franchi, ``{6D} physical interaction
  with an omnidirectional aerial robot,'' \emph{Drones}, vol.~10, no.~2, p.
  129, 2026.

\bibitem{Xu2023PhysicalSurvey}
J.~Xu, ``Multi-rotor aerial vehicles in physical interactions: A survey,''
  \emph{arXiv preprint arXiv:2312.02635}, 2023.

\bibitem{Brandt2011Propeller}
J.~B. Brandt and M.~S. Selig, ``Propeller performance data at low reynolds
  numbers,'' in \emph{49th AIAA Aerospace Sciences Meeting including the New
  Horizons Forum and Aerospace Exposition}, 2011.

\bibitem{McCrink2017BladeElement}
M.~H. McCrink and J.~W. Gregory, ``Blade element momentum modeling of
  low-reynolds electric propulsion systems,'' \emph{Journal of Aircraft},
  vol.~54, no.~1, pp. 163--176, 2017.

\bibitem{Mahony2012Multirotor}
R.~Mahony, V.~Kumar, and P.~Corke, ``Multirotor aerial vehicles: Modeling,
  estimation, and control of quadrotor,'' \emph{IEEE Robotics \& Automation
  Magazine}, vol.~19, no.~3, pp. 20--32, 2012.

\bibitem{Davoudi2020QuadRotor}
B.~Davoudi, E.~Taheri, K.~Duraisamy, B.~Jayaraman, and I.~Kolmanovsky,
  ``Quad-rotor flight simulation in realistic atmospheric conditions,''
  \emph{AIAA Journal}, vol.~58, no.~5, pp. 1992--2004, 2020.

\bibitem{Bangura2014Aerodynamic}
M.~Bangura, H.~Lim, H.~J. Kim, and R.~Mahony, ``Aerodynamic power control for
  multirotor aerial vehicles,'' in \emph{2014 IEEE International Conference on
  Robotics and Automation (ICRA)}, 2014, pp. 529--536.

\bibitem{Aboudorra2024OmniMorph}
Y.~Aboudorra, C.~Gabellieri, R.~Brantjes, Q.~Sabl{\'e}, and A.~Franchi,
  ``Modelling, analysis, and control of {OmniMorph}: An omnidirectional
  morphing multi-rotor {UAV},'' \emph{Journal of Intelligent \& Robotic
  Systems}, vol. 110, no.~1, p.~21, 2024.

\bibitem{Mohamed2011Designing}
M.~K. Mohamed, S.~Patra, and A.~Lanzon, ``Designing electric propulsion systems
  for {UAVs},'' in \emph{Towards Autonomous Robotic Systems}, ser. Lecture
  Notes in Computer Science.\hskip 1em plus 0.5em minus 0.4em\relax Springer,
  2011, vol. 6856, pp. 388--389.

\bibitem{Gatti2017Complete}
M.~Gatti, ``Complete preliminary design methodology for electric multirotor,''
  \emph{Journal of Aerospace Engineering}, vol.~30, no.~5, p. 04017046, 2017.

\bibitem{Alulema2021Sizing}
V.~Alulema, E.~Valencia, E.~Cando, V.~Hidalgo, and D.~Rodriguez, ``Propulsion
  sizing correlations for electrical and fuel powered unmanned aerial
  vehicles,'' \emph{Aerospace}, vol.~8, no.~7, p. 171, 2021.

\bibitem{Dai2026HighPrecision}
X.~Dai, W.~Liu, and Y.~Chen, ``High-precision modeling of {UAV} electric
  propulsion for improving endurance estimation,'' \emph{Drones}, vol.~10,
  no.~2, p.~80, 2026.

\bibitem{Traub2011Range}
L.~W. Traub, ``Range and endurance estimates for battery-powered aircraft,''
  \emph{Journal of Aircraft}, vol.~48, no.~2, pp. 703--707, 2011.

\bibitem{Paek2020Battery}
S.~W. Paek, S.~Kim, and R.~Vinoth, ``Optimal endurance and range of electric
  aircraft with battery degradation,'' \emph{Transactions of the Japan Society
  for Aeronautical and Space Sciences}, vol.~63, no.~2, pp. 62--65, 2020.

\bibitem{Hwang2018Endurance}
M.-H. Hwang, H.-R. Cha, and S.~Y. Jung, ``Practical endurance estimation for
  minimizing energy consumption of multirotor unmanned aerial vehicles,''
  \emph{Energies}, vol.~11, no.~9, p. 2221, 2018.

\bibitem{Hassanalian2014Solar}
M.~Hassanalian, M.~Radmanesh, and A.~Sedaghat, ``Increasing flight endurance of
  {MAVs} using multiple quantum well solar cells,'' \emph{International Journal
  of Aeronautical and Space Sciences}, vol.~15, no.~2, pp. 212--217, 2014.

\bibitem{Abeywickrama2018Comprehensive}
H.~V. Abeywickrama, B.~A. Jayawickrama, Y.~He, and E.~Dutkiewicz,
  ``Comprehensive energy consumption model for unmanned aerial vehicles, based
  on empirical studies of battery performance,'' \emph{IEEE Access}, vol.~6,
  pp. 58\,383--58\,394, 2018.

\bibitem{Kirschstein2020Delivery}
T.~Kirschstein, ``Comparison of energy demands of drone-based and ground-based
  parcel delivery services,'' \emph{Transportation Research Part D: Transport
  and Environment}, vol.~78, p. 102209, 2020.

\bibitem{Beigi2022Overview}
P.~Beigi, M.~S. Rajabi, and S.~Aghakhani, ``An overview of drone energy
  consumption factors and models.''\hskip 1em plus 0.5em minus 0.4em\relax
  Springer, 2022.

\bibitem{Michel2023EnergyOptimal}
N.~Michel, P.~Wei, Z.~Kong, and X.~Lin, ``Energy-optimal unmanned aerial
  vehicles motion planning and control based on integrated system physical
  dynamics,'' \emph{Journal of Dynamic Systems, Measurement, and Control}, vol.
  145, no.~4, p. 041002, 2023.

\bibitem{DiFranco2016Coverage}
C.~D. Franco and G.~Buttazzo, ``Coverage path planning for {UAVs}
  photogrammetry with energy and resolution constraints,'' \emph{Journal of
  Intelligent \& Robotic Systems}, vol.~83, pp. 445--462, 2016.

\bibitem{Datsko2024Coverage}
D.~Datsko, F.~Nekovar, R.~Penicka, and M.~Saska, ``Energy-aware multi-{UAV}
  coverage mission planning with optimal speed of flight,'' \emph{IEEE Robotics
  and Automation Letters}, vol.~9, no.~3, pp. 2893--2900, 2024.

\bibitem{Wu2025PathPlanning}
K.~Wu, J.~Lan, S.~Lu, C.~Wu, B.~Liu, and Z.~Lu, ``Integrative path planning for
  multi-rotor logistics {UAVs} considering {UAV} dynamics, energy efficiency,
  and obstacle avoidance,'' \emph{Drones}, vol.~9, no.~2, p.~93, 2025.

\bibitem{Zeng2019Energy}
Y.~Zeng, J.~Xu, and R.~Zhang, ``Energy minimization for wireless communication
  with rotary-wing {UAV},'' \emph{IEEE Transactions on Wireless
  Communications}, vol.~18, no.~4, pp. 2329--2345, 2019.

\bibitem{Ribeiro2024RotaryFixed}
P.~Ribeiro, A.~Coelho, and R.~Campos, ``On the energy consumption of rotary
  wing and fixed wing {UAVs} in flying networks,'' \emph{arXiv preprint
  arXiv:2406.19009}, 2024.

\bibitem{Bronz2009Endurance}
M.~Bronz, J.-M. Moschetta, P.~Brisset, and M.~Gorraz, ``Towards a long
  endurance {MAV},'' \emph{International Journal of Micro Air Vehicles},
  vol.~1, no.~4, pp. 241--254, 2009.

\bibitem{DeCroon2009DelFly}
G.~C. H.~E. de~Croon, K.~M.~E. de~Clercq, R.~Ruijsink, B.~D.~W. Remes, and
  C.~de~Wagter, ``Design, aerodynamics, and vision-based control of the
  {DelFly},'' \emph{International Journal of Micro Air Vehicles}, vol.~1,
  no.~2, pp. 71--97, 2009.

\bibitem{Karasek2018Flapper}
M.~Kar{\'a}sek, F.~T. Muijres, C.~de~Wagter, B.~D.~W. Remes, and G.~C. H.~E.
  de~Croon, ``A tailless aerial robotic flapper reveals that flies use torque
  coupling in rapid banked turns,'' \emph{Science}, vol. 361, no. 6407, pp.
  1089--1094, 2018.

\bibitem{Smeur2020Incremental}
E.~J.~J. Smeur, M.~Bronz, and G.~C. H.~E. de~Croon, ``Incremental control and
  guidance of hybrid aircraft applied to a tailsitter unmanned air vehicle,''
  \emph{Journal of Guidance, Control, and Dynamics}, vol.~43, no.~2, pp.
  274--287, 2020.

\bibitem{Mousaei2022TiltRotor}
M.~Mousaei, J.~Geng, A.~Keipour, D.~Bai, and S.~Scherer, ``Design, modeling and
  control for a tilt-rotor {VTOL UAV} in the presence of actuator failure,'' in
  \emph{2022 IEEE/RSJ International Conference on Intelligent Robots and
  Systems (IROS)}, 2022, pp. 4310--4317.

\bibitem{Ryll2015Overactuated}
M.~Ryll, H.~H. B{\"u}lthoff, and P.~R. Giordano, ``A novel overactuated
  quadrotor unmanned aerial vehicle: Modeling, control, and experimental
  validation,'' \emph{IEEE Transactions on Control Systems Technology},
  vol.~23, no.~2, pp. 540--556, 2015.

\bibitem{SanchezCuevas2017Ground}
P.~J. Sanchez-Cuevas, G.~Heredia, and A.~Ollero, ``Characterization of the
  aerodynamic ground effect and its influence in multirotor control,''
  \emph{International Journal of Aerospace Engineering}, vol. 2017, p. 1823056,
  2017.

\bibitem{SanchezCuevas2020Obstacles}
P.~J. Sanchez-Cuevas, V.~Martin, G.~Heredia, and A.~Ollero, ``Aerodynamic
  effects in multirotors flying close to obstacles: Modelling and mapping,'' in
  \emph{Robot 2019: Fourth Iberian Robotics Conference}.\hskip 1em plus 0.5em
  minus 0.4em\relax Springer, 2020, pp. 63--74.

\end{thebibliography}
\end{document}